\documentclass[10pt,twocolumn,letterpaper]{article}

\usepackage{cvpr}              
\usepackage{graphicx}
\usepackage{caption}
\usepackage{xcolor}
\usepackage{enumitem}
\usepackage{amssymb}
\usepackage{bbm}
\usepackage{mathtools}
\usepackage{mathabx}
\usepackage{comment}
\usepackage{algorithm} 
\usepackage{algpseudocode} 

\definecolor{cvprblue}{rgb}{0.21,0.49,0.74}
\usepackage[pagebackref,breaklinks,colorlinks,citecolor=cvprblue]{hyperref}

\def\paperID{14571}
\def\confName{CVPR}
\def\confYear{2024}

\newcommand\given[1][]{\:#1\vert\:}

\DeclareMathOperator{\data}{data}
\newcommand{\innerproduct}[2]{\langle #1, #2 \rangle}
\newcommand\norm[1]{\left\lVert#1\right\rVert}

\newcommand{\xb}{{\boldsymbol x}}

\newcommand{\yb}{{\boldsymbol y}}

\newcommand{\vb}{{\boldsymbol v}}

\newcommand{\epsilonb}{{\boldsymbol \epsilon}}

\newcommand{\Eb}{{\mathbb E}}

\newcommand{\alphabar}{{\bar \alpha}}

\usepackage{amsthm}

\newcommand{\Ib}{{\boldsymbol I}}

\newcommand{\x}{{\boldsymbol x}}

\renewcommand{\t}{{\boldsymbol t}}

\newcommand{\Nc}{{\mathcal N}}

\newcommand{\Pc}{{\mathcal P}}

\newcommand\Mark[1]{\textsuperscript#1}

\title{VMC: Video Motion Customization using Temporal Attention Adaption for Text-to-Video Diffusion Models}

\author{
Hyeonho Jeong\Mark{1}\thanks{equal} \qquad 
Geon Yeong Park\Mark{2}\footnotemark[1] \qquad 
Jong Chul Ye\Mark{1}\Mark{,}\Mark{2}\\
\Mark{1}Kim Jaechul Graduate School of AI, 
\Mark{2}Bio and Brain Engineering\\
Korea Advanced Institute of Science and Technology (KAIST)\\
* indicates co-first authors\\
\texttt{\{hyeonho.jeong, pky3436, jong.ye\}@kaist.ac.kr} 
}

\begin{document}

\twocolumn[{%
\renewcommand\twocolumn[1][]{#1}%
\maketitle
\begin{center}
    \centering
    \captionsetup{type=figure}
    \includegraphics[width=0.96\textwidth]{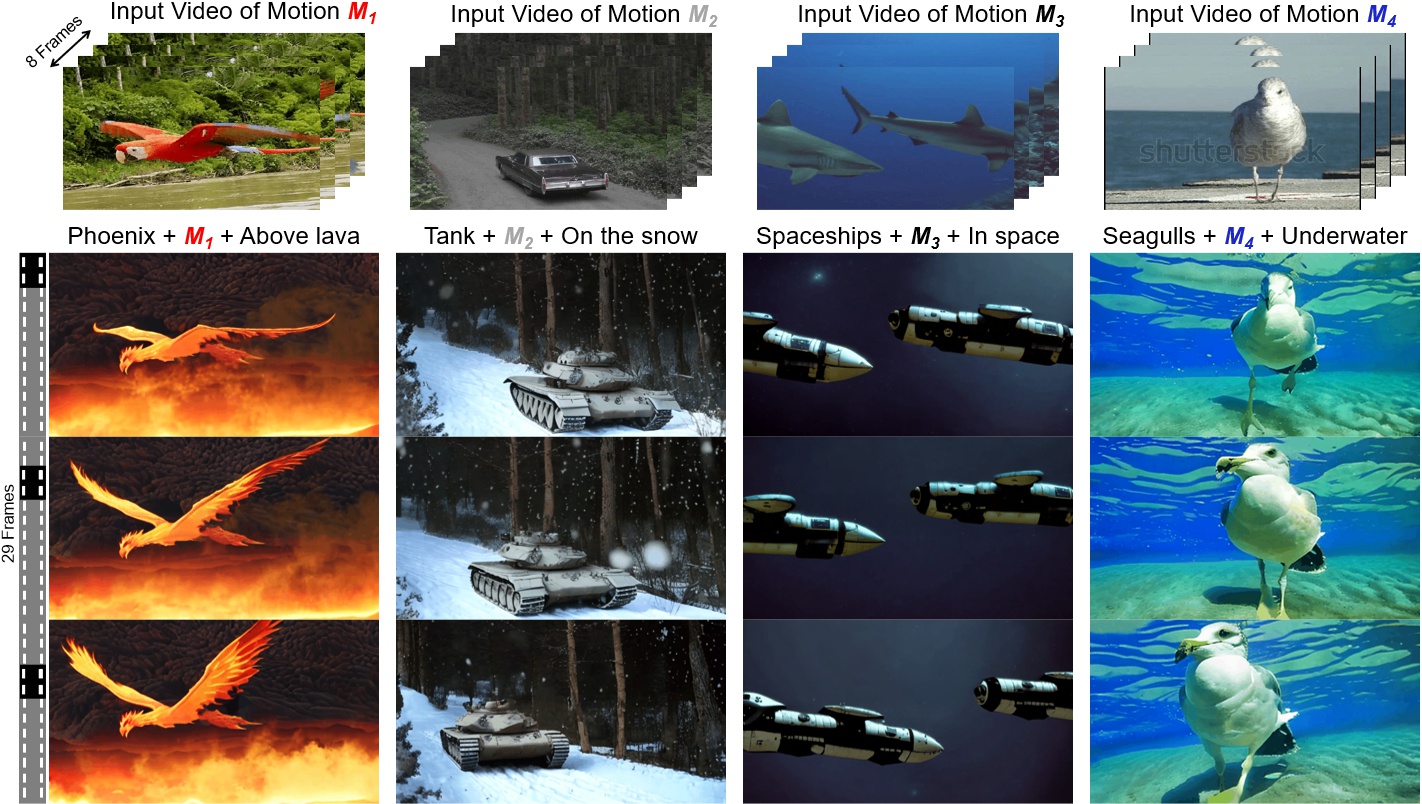}
    \captionof{figure}{{Using only a single video portraying any type of motion, our \textbf{V}ideo \textbf{M}otion \textbf{C}ustomization framework allows for generating a wide variety of videos characterized by the same motion but in entirely distinct contexts and better spatial/temporal resolution.} 8-frame input videos are translated
    to 29-frame videos in different contexts while closely following the target motion. The visualized frames for the first video are at indexes 1, 9, and 17. A comprehensive view of these motions in the form of videos can be explored at our 
    \href{https://video-motion-customization.github.io/}{project page}.
    }
    \label{fig-teaser}
\end{center}%
}]

\begin{abstract}
Text-to-video diffusion models have advanced video generation significantly. However,
customizing these models to generate videos with tailored motions presents a substantial challenge. 
In specific,
they encounter hurdles in (\textbf{a}) accurately reproducing motion from a target video, and (\textbf{b}) creating diverse visual variations. 
For example, straightforward extensions of static image customization methods to video often lead to intricate entanglements of appearance and motion data. To tackle this, here we present the Video Motion Customization (\textbf{VMC}) framework, a novel one-shot tuning approach crafted to adapt temporal attention layers within video diffusion models. 
Our approach introduces a novel motion distillation objective using residual vectors between consecutive frames as a motion reference. The diffusion process then 
preserves low-frequency motion trajectories while mitigating high-frequency motion-unrelated noise in image space. We validate our method against state-of-the-art video generative models across diverse real-world motions and contexts. 
Our codes, data and the project demo can be found at \href{https://video-motion-customization.github.io/}
{https://video-motion-customization.github.io}.


\end{abstract}

\section{Introduction}
\quad 
The evolution of diffusion models \cite{sohl2015deep, ho2020denoising, song2020score} has significantly advanced Text-to-Image (T2I) generation, notably when paired with extensive text-image datasets \cite{schuhmann2022laion, kakaobrain2022coyo-700m}. While cascaded diffusion pipelines \cite{ho2022imagen, singer2022make, zhou2022magicvideo, blattmann2023align, ge2023preserve, wang2023lavie, zhang2023show} have extended this success to Text-to-Video (T2V) generation, current models lack the ability to replicate specific motions or generate diverse variations of the same motion with distinct visual attributes and backgrounds. Addressing this, we tackle the challenge of Motion Customization \cite{zhao2023motiondirector}—adapting pre-trained Video Diffusion Models (VDM) to produce motion-specific videos
 in different contexts, while maintaining the same motion patterns of target subjects.

Given a few subject images for reference, appearance customization \cite{gal2022image, ruiz2023dreambooth, shi2023instantbooth, ruiz2023hyperdreambooth, ma2023subject, wu2023tune} in generative models aims to fine-tune models to generate subject images in diverse contexts. However, these approaches, despite varying optimization objectives, commonly strive for \textit{faithful} image (frame) reconstruction by minimizing the $\ell_2$-distance between predicted and ground-truth noise. This may lead to the \textit{entangled} learning of appearance and motion.

To tackle this, we present \textbf{VMC}, a new framework aimed at adapting pre-trained VDM's temporal attention layers via our proposed \textit{Motion Distillation} objective. This approach utilizes residual vectors between consecutive (latent) frames to obtain the motion vectors that trace motion trajectories in the target video.
 Consequently, we fine-tune VDM's temporal attention layers to align the ground-truth image-space residuals with their denoised estimates, which equivalently aligns predicted and ground-truth source noise differences within VDM.
 This enables lightweight and fast one-shot training. 
To further facilitate the appearance-invariant motion distillation, we transform faithful text prompts into appearance-invariant prompts, e.g. \texttt{"A bird is flying above a lake in the forest"} $\rightarrow$ \texttt{"A bird is flying"} in Fig. \ref{fig-teaser}. This encourages the modules to focus on the motion information and ignore others, such as appearance, distortions, background, etc. 
During inference, our procedure initiates by sampling key-frames using the adapted key-frame generation U-Net, followed by temporal interpolation and spatial super-resolution. 
To summarize, VMC makes the following key contributions:

\begin{itemize}[noitemsep]
    \item We introduce a novel fine-tuning strategy which focuses solely on temporal attention layers in the key-frame generation module. This enables lightweight training (15GB vRAM) and fast training ($<$ 5 minutes). 
    \item To our knowledge, we mark a pioneering case of fine-tuning only the temporal attention layers in video diffusion models, without optimizing spatial self or cross-attention layers, while achieving successful motion customization.
    \item We introduce a novel motion distillation objective that leverages the residual vectors between consecutive (latent) frames as motion vectors. 
    \item We present the concept of appearance-invariant prompts, which further facilitates the process of motion learning when combined with our motion distillation loss.
\end{itemize}

\begin{figure*}[htbp]
    \centering
    \includegraphics[width=\textwidth]
    {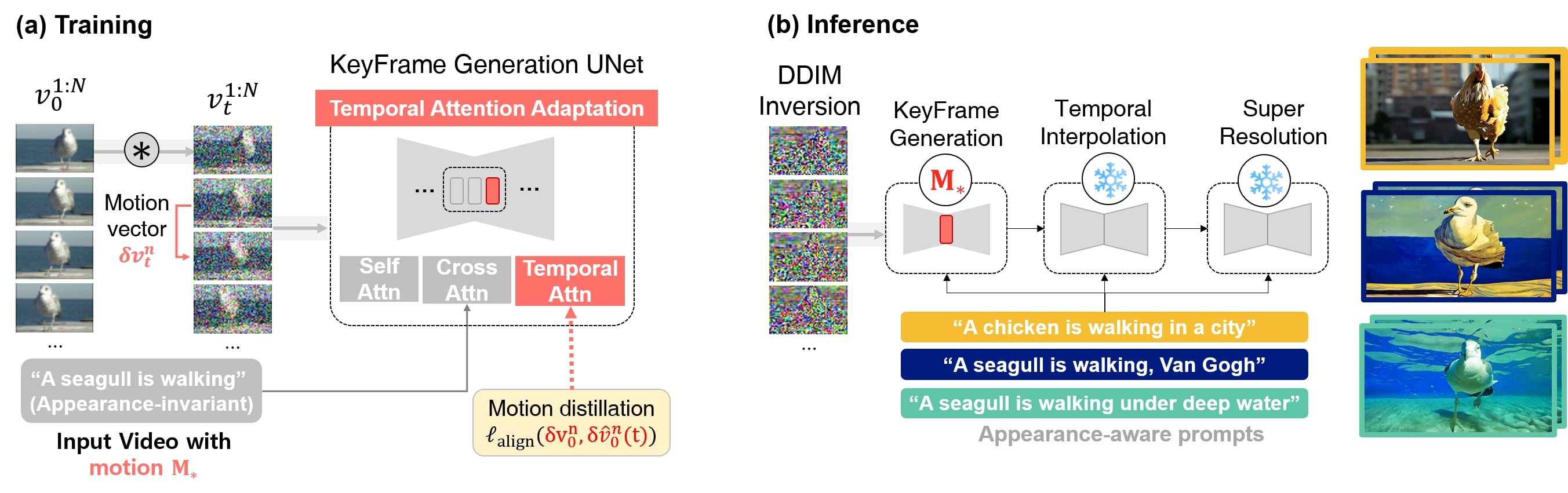}
    \caption{\textbf{Overview}. The proposed Video Motion Customization (\textbf{VMC}) framework distills the motion trajectories from the residual between consecutive (latent) frames, namely motion vector $\delta \vb_t^n$ for $\t \geq 0$. We fine-tune only the temporal attention layers of the key-frame generation model by aligning the ground-truth and predicted motion vectors. After training, the customized key-frame generator is leveraged for target motion-driven video generation with new appearances context, e.g. \texttt{"A chicken is walking in a city"}.}
    \label{fig-overview}
\end{figure*}

\section{Preliminaries}
\label{sec: prelim}
\paragraph{Diffusion Models.} Diffusion models aim to generate samples from the Gaussian noise through iterative denoising processes. Given a clean sample $\xb_0 \sim p_{\data}(\xb)$, the forward process is defined as a Markov chain with forward conditional densities
\begin{equation}
    \label{eq: diffusion forward density}
    \begin{split}
        p(\xb_t \given \xb_{t-1}) = \Nc (\xb_t \given \beta_t \xb_{t-1}, (1-\beta_t) I) \\
        p_t(\xb_t \given[] \xb_0) = \Nc (\xb_t \given \sqrt{\alphabar} \xb_0, (1-\alphabar) I),
    \end{split} 
\end{equation}
where $\xb_{t} \in \mathbb{R}^d$ is a noisy latent variable at a timestep $t$ that has the same dimension as $\xb_0$, and $\beta_t$ denotes an increasing sequence of noise schedule where $\alpha_t \coloneqq 1 - \beta_t$ and $\alphabar_t \coloneqq \Pi_{i=1}^t \alpha_i$. Then, the goal of diffusion model training is to obtain a residual denoiser $\epsilonb_{\theta}$: 
\begin{equation}
    \label{eq: epsilon matching}
    \min_{\theta} \Eb_{\xb_t \sim p_t(\xb_t \given \xb_0), \xb_0 \sim p_{\data}(\xb_0), \epsilonb \sim \Nc(0, I)} \big[ \norm{\epsilonb_{\theta} (\xb_t, t) - \epsilonb} \big] .
\end{equation}
It can be shown that this epsilon matching in \eqref{eq: epsilon matching} is equivalent to the Denoising Score Matching (DSM  \cite{hyvarinen2005estimation, song2019generative}) with different parameterization:
\begin{equation}
    \label{eq: score matching}
    \min_{\theta} \Eb_{\xb_t, \xb_0, \epsilonb} \big[ \norm{ \boldsymbol{s}_{\theta}^t (\xb_t) - \nabla_{\xb_t} \log p_t(\xb_t \given \xb_0) } \big],
\end{equation}
where $\boldsymbol{s}_{\theta^*}(\xb_t, t) \simeq - \frac{\xb_t - \sqrt{\alphabar_t} \xb_0 }{1-\alphabar} = - \frac{1}{\sqrt{1-\alphabar_t}} \epsilonb_{\theta^*} (\xb_t, t)$. The reverse sampling from $q(\xb_{t-1}|\xb_t, \epsilonb_{\theta^*}(\xb_t, t))$ is then achieved by
\begin{align}
    \label{eq: reverse sampling}
    \x_{t-1}=\frac{1}{\sqrt{\alpha_t}}\Big{(}\x_t-\frac{1-\alpha_t}{\sqrt{1-\bar{\alpha}_t}}\boldsymbol{\epsilon}_{\theta^*}(\x_t, t)\Big{)}+\tilde{\beta}_t \epsilonb,
\end{align}
where $\epsilonb \sim\Nc(0,\Ib)$ and $\tilde{\beta}_t \coloneqq \frac{1 - \alphabar_{t-1}}{1 - \alphabar_t} \beta_t$. 
To accelerate sampling, DDIM \cite{song2020denoising} further proposes another sampling method as follows:
\begin{equation}
    \label{eq: ddim sampling}
    \xb_{t-1} = \sqrt{\alphabar_{t-1}} \hat{\xb}_0(t) + \sqrt{ 1 - \alphabar_{t-1} - \eta^2 \tilde{\beta_t}^2} \epsilonb_{\theta^*} (\xb_t, t) + \eta \tilde{\beta}_t \epsilonb,
\end{equation}
where $\eta \in [0,1]$ is a stochasticity parameter, and $\hat{\xb}_0(t)$ is the denoised estimate which can be equivalently derived using Tweedie's formula \cite{efron2011tweedie}:
\begin{equation}
    \label{eq: Tweedie}
    \hat{\xb}_0(t) \coloneqq \frac{1}{\sqrt{\alphabar_t}} (\xb_t - \sqrt{1-\alphabar_t} \epsilonb_{\theta^*} (\xb_t, t)).
\end{equation}
For a text-guided Diffusion Model, the training objective is often given by:
\begin{equation}
    \label{eq: condition epsilon matching}
    \min_{\theta} \Eb_{\xb_t, \xb_0, \epsilonb, c} \big[ \norm{\epsilonb_{\theta} (\xb_t, t, c) - \epsilonb} \big],
\end{equation}
where $c$ represents the textual embedding. Throughout this paper, we will often omit $c$ from $\epsilonb_{\theta} (\xb_t, t, c)$ if it does not lead to notational ambiguity.


\paragraph{Video Diffusion Models.} Video diffusion models \cite{ho2022imagen, zhang2023show, ho2204video} further attempt to model the video data distribution. Specifically, Let $(\vb^n)_{n \in \{1, \dots, N\}}$ represents the $N$-frame input video sequence. Then, for a given $n$-th frame $\vb^n \in \mathbb{R}^d$, let $\vb^{1:N} \in \mathbb{R}^{N \times d}$ represents a whole video vector. Let $\boldsymbol{v}_t^n = \sqrt{\bar{\alpha}_t} \boldsymbol{v}^n + \sqrt{1 - \bar{\alpha}_t} \boldsymbol{\epsilon}_t^{n}$ represents the $n$-th noisy frame latent sampled from $p_t(\boldsymbol{v}_t^n | \boldsymbol{v}^n)$, where $\boldsymbol{\epsilon}_t^n \sim \mathcal{N}(0, I)$. We similarly define $(\vb_t^n)_{n \in 1, \dots, N}$, $\vb_t^{1:N}$, and $\epsilonb_t^{1:N}$. The goal of video diffusion model training is then to obtain a residual denoiser $\epsilonb_{\theta} $ with textual condition $c$ and video input that satisfies:
\begin{equation}
    \label{eq: video epsilon matching}
    \min_{\theta} \Eb_{\vb_t^{1:N}, \vb^{1:N}, \epsilonb_t^{1:N}, c} \big[ \norm{\epsilonb_{\theta} (\vb_t^{1:N}, t, c) - \epsilonb_t^{1:N}} \big],
\end{equation}
where $\epsilonb_\theta (\vb_t^{1:N}, t, c), \epsilonb_t^{1:N} \in \mathbb{R}^{N \times d}$. In this work, we denote the predicted noise of $n$-th frame as $\epsilonb_\theta^n (\vb_t^{1:N}, t, c) \in \mathbb{R}^d$.

In practice, contemporary video diffusion models often employ cascaded inference pipelines for high-resolution outputs. For instance, \cite{zhang2023show} initially generates a low-resolution video with strong text-video correlation, further enhancing its resolution via temporal interpolation and spatial super-resolution modules.

In exploring video generative tasks through diffusion models, two primary approaches have emerged: foundational Video Diffusion Models (VDMs) or leveraging pre-trained Text-to-Image (T2I) models. To extend image diffusion models to videos, several architectural modifications are made. Typically, U-Net generative modules integrate temporal attention blocks after spatial attentions \cite{ho2204video}. Moreover, 2D convolution layers are inflated to 3D convolution layers by altering kernels \cite{ho2204video}.  

\section{Video Motion Customization}
\quad Given an input video, our main goal is to (\textbf{a}) distill the motion patterns $M_*$ of target subjects, and (\textbf{b}) customize the input video in different contexts while maintaining the same motion patterns $M_*$, e.g. \texttt{Sharks w/ motion $M_*$} $\rightarrow$ \texttt{Airplanes w/ motion $M_*$}, with minimal computational costs. 

To this end, we propose a novel video motion customization framework, namely \textbf{VMC}, which leverages cascaded video diffusion models with robust temporal priors. One notable aspect of the proposed framework is that we perform fine-tuning \textit{only} on the key-frame generation module, also referred to as the T2V base model, within the cascaded VDMs, which guarantees computational and memory efficiency. Specifically, within the key-frame generation model, our fine-tuning process \textit{only} targets the temporal attention layers. This facilitates adaptation while preserving the model's inherent capacity for generic synthesis. Notably, we \textit{freeze} the subsequent frame interpolation and spatial super-resolution modules as-is (Fig. \ref{fig-overview}).

\subsection{Temporal Attention Adaptation}
\label{sec: temporal adaptation}
\quad In order to distill the motion $M_*$, we first propose a new objective function for temporal attention adaptation using residual cosine similarity. Our intuition is that residual vectors between consecutive frames may include information about the motion trajectories. 

Let $(\vb^n)_{n \in \{1, \dots, N\}}$ represents the $N$-frame input video sequence. As defined in Section \ref{sec: prelim}, for a given noisy video latent vector $\boldsymbol{v}_t^{1:N}$ with $\epsilonb_t^{1:N}$, let $\boldsymbol{v}_t^n$ represents the $n$-th noisy frame latent sampled from $p_t(\boldsymbol{v}_t^n | \boldsymbol{v}^n)$ with $\epsilonb_t^n$. We will interchangeably use $\vb^n$ and $\vb_0^n$ for notational simplicity. Likewise, $\boldsymbol{v}_t^{n+c}$ is defined as $\boldsymbol{v}_t^n$, with $c > 0$ representing the fixed frame stride. Then, we define the frame residual vector at time $t \geq 0$ as 
\begin{equation}
    \label{eq: residual frame}
    \delta \vb_t^{n} \coloneqq \vb_t^{n+c} - \vb_t^{n},
\end{equation}
where we similarly define the epsilon residual vector $\delta \epsilonb_t^n$. In the rest of the paper, we interchangeably use frame residual vector and \textit{motion vector}.

We expect that these motion vectors may encode information about motion patterns, where such information may vary depending on the time $t$ and its corresponding noise level. The difference vector $\delta \vb_t^{n}$ can be delineated as:
\begin{equation}
    \label{eq: residual frame decomposition}
    \begin{split}
        \delta \vb_t^n &= \sqrt{\alphabar_t} (\vb_0^{n+c} - \vb_0^{n}) + \sqrt{1 - \alphabar_t} (\epsilonb_t^{n+c} -\epsilonb_t^n) \\
        &=\sqrt{\alphabar_t} \delta \vb_0^n + \sqrt{ 1 - \alphabar_t} \delta \epsilonb_t^n,  
    \end{split}
\end{equation}
where $\delta \epsilonb_t^n$ is normally distributed with zero mean and $2I$ variance. In essence, $\delta \vb_t^n$ can be acquired through the following diffusion kernel:
\begin{equation}
    \label{eq: residual frame kernel}
    p(\delta \vb_t^n \given \delta \vb_0^n) = \Nc (\delta \vb_t^n \given \sqrt{\alphabar_t} \delta \vb_0^{n}, 2(1 - \alphabar_t)I).
\end{equation}

\begin{figure}
    \centering
    \vspace{-5pt}
    \includegraphics[width=\linewidth]{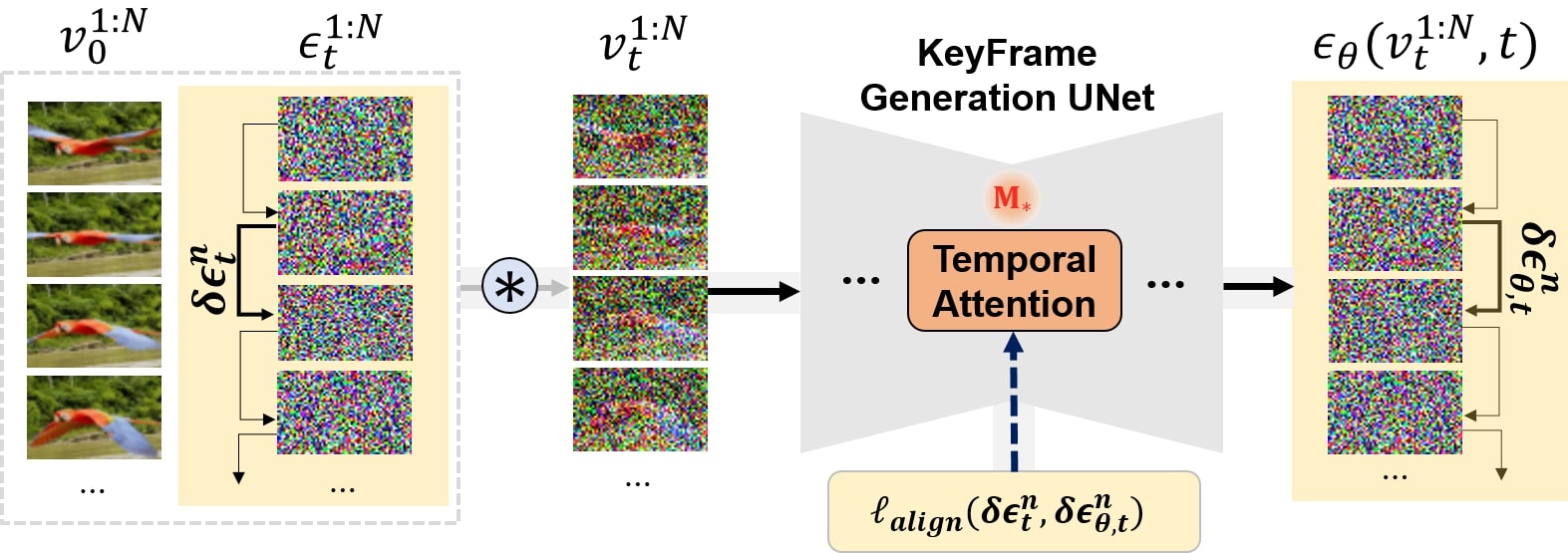}
    \caption{\textbf{Training.} The proposed framework aims to learn motion by $\delta \epsilonb_t^n$-alignment using  \eqref{eq: mse} or \eqref{eq: optimization framework}. Note that we only fine-tune the temporal attention layers in the key-frame generation U-Net. The blue circle represents the diffusion forward process.}
    \label{fig-model}
\end{figure}

In light of this, our goal is to transfer motion information to the temporal attention layers by leveraging the motion vectors. For this, we first simulate the motion vectors using video diffusion models. 
{Specifically, as similarly done in \eqref{eq: Tweedie}, the denoised video vector estimates $\hat{\vb}_0^{1:N}(t)$ can be derived by applying Tweedie's formula:
\begin{equation}
    \label{eq: video Tweedie}
    \hat{\vb}_0^{1:N}(t) \coloneqq \frac{1}{\sqrt{\alphabar_t}} \big( \vb_t^{1:N} - \sqrt{1-\alphabar_t} \epsilonb_{\theta} (\vb_t^{1:N}, t) \big),
\end{equation}
where $\hat{\vb}_0^{1:N}(t)$ is an empirical Bayes optimal posterior expectation $\mathbb{E} [\vb_0^{1:N} \given \vb_t^{1:N}]$.}
Then, the denoised motion vector estimate $\delta \hat{\vb}_0^n$ can be defined in terms of $\delta \vb_t^n$ and $\delta \epsilonb_\theta^n (\vb_t^{1:N}, t)$ by using \eqref{eq: video Tweedie}: 
\begin{equation}
    \label{eq: frame residual Tweedie estimate}
    \delta \hat{\vb}_0^n(t) \coloneqq \frac{1}{\sqrt{\alphabar_t}} \Big( \delta \vb_t^n - \sqrt{1-\alphabar_t} \delta \epsilonb_{\theta, t}^n \Big),
\end{equation}
where $\delta \epsilonb_\theta^n(\vb_t^{1:N}, t) \coloneqq \epsilonb_\theta^{n+c}(\vb_t^{1:N}, t) - \epsilonb_\theta^n(\vb_t^{1:N}, t)$ is abbreviated as $\delta \epsilonb_{\theta, t}^n$ for notational simplicity.
Similarly, one can obtain ground-truth motion vector $\delta \vb_0^n$ in terms of $\delta \vb_t^n$ and $\delta \epsilonb_t^n$ by using \eqref{eq: residual frame decomposition}:
\begin{equation}
    \label{eq: frame residual Tweedie ground-truth}
    \delta \vb_0^n = \frac{1}{\sqrt{\alphabar_t}} \Big( \delta \vb_t^n - \sqrt{1-\alphabar_t} \delta \epsilonb_t^n \Big).
\end{equation}

Then, our objective is to finetune $\theta$ by \textit{aligning} the motion vector $\delta \vb_0^n$ and its denoised estimate $\delta \hat{\vb}_0^n(t)$:
\begin{equation}
    \label{eq: v0 align}
    \min_{\theta} \mathbb{E}_{t, n, \epsilonb_t^n, \epsilonb_t^{n+c}} \Big[\ell_{\text{align}} \big(\delta \vb_0^n, \delta \hat{\vb}_0^n(t) \big) \Big],
\end{equation}
with a loss function $\ell_{\text{align}}: \mathbb{R}^d \times \mathbb{R}^d \rightarrow \mathbb{R}$. By using $\ell_2$-distance for $\ell_{\text{align}}$,  this is equivalent to matching $\delta \epsilonb_{\theta, t}^n$ and $\delta \epsilonb_t^n$:
\begin{equation}
    \begin{split}
        \label{eq: mse}
        \ell_{\text{align}} \big(\delta \vb_0^n, \delta \hat{\vb}_0^n(t) \big) = \frac{1-\alphabar_t}{\alphabar_t} \norm{ \delta \epsilonb_t^n - \delta \epsilonb_{\theta,t}^n }^2. \\
    \end{split}
\end{equation}
Notably, aligning the ground-truth and predicted motion vectors translates into aligning epsilon residuals. 

While this objective demonstrates effective empirical performance, our additional observations indicate that using $\ell_{\text{cos}} (\delta \epsilonb_t^n, \delta \epsilonb_{\theta, t}^n)$ may further improve the distillation, where $\ell_{\text{cos}}(\xb, \yb) = 1 - \frac{ \innerproduct{\xb}{\yb} }{\|\xb\| \|\yb\|}$ for $\xb, \yb \in \mathbb{R}^d$ (more analysis in section \ref{sec: ablation}). 
Accordingly, our optimization framework is finally defined as follows:
\begin{equation}
    \label{eq: optimization framework}
        \min_{\theta} \mathbb{E}_{t, n, \epsilonb_t^n, \epsilonb_t^{n+c}} [ \ell_{\text{cos}} (\delta \epsilonb_t^n, \delta \epsilonb_{\theta, t}^n) ].
\end{equation}
Thus, the proposed optimization framework aims to \textit{maximize} the residual cosine similarity between $\delta \epsilonb_t^n$ and $\delta \epsilonb_{\theta,t}^n$. In our observation, aligning the image-space residuals ($\delta \vb_0^n$ and $\delta \hat{\vb}_0^n(t)$) corresponds to aligning the latent-space epsilon residuals ($\delta \epsilonb_t^n$ and $\delta \epsilonb_{\theta, t}^n$) across varying time steps. This relationship stems from expressing the motion vector $\delta \vb_0^n$ and its estimation $\delta \hat{\vb}_0^n(t)$ in terms of $\delta \vb_t^n$, $\delta \epsilonb_t^n$, and $\delta \epsilonb_{\theta, t}^n$. Consequently, the proposed optimization framework fine-tunes temporal attention layers by leveraging diverse diffusion latent spaces at time $t$ which potentially contains multi-scale rich descriptions of video frames. Hence, this optimization approach can be seamlessly applied to video diffusion models trained using epsilon-matching, thanks to the equivalence between $\delta \epsilonb_t^n$-matching and $\delta \vb_0^n$-matching. Practically, we exclusively fine-tune the temporal attention layers $\theta_{\text{TA}} \subset \theta$, originally designed for dynamic temporal data assimilation \cite{wu2023tune}. The frame stride remains fixed at $c=1$ across all experiments.

\subsection{Appearance-invariant Prompts}
\label{method-prompting}
\begin{figure}[!t]
    \centering
    \vspace{-5pt}
    \includegraphics[width=\columnwidth]
    {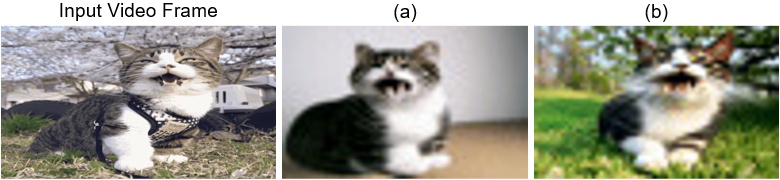}
    \caption{\textbf{Appearance-invariant Prompt}. Comparison of input reconstruction with and without appearance-invariant prompt:
    (a) and (b) depict sampled low-resolution (64x40) keyframes.
    For (a), the training prompt used was ``A cat is roaring," while for (b), the training prompt was ``A cat is roaring on the grass under the tree." Our appearance-invariant prompt enables the removal of background information that can disturb
    motion distillation.  
}
    \label{fig-api}
\end{figure}

\quad In motion distillation, it is crucial to filter out disruptive variations that are unrelated to motion. These variations may include changes in appearance and background, distortions, consecutive frame inconsistencies, etc. To achieve this, we further utilize \textit{appearance-invariant prompts}. 
Diverging from traditional generative customization frameworks \cite{ruiz2023dreambooth, ruiz2023hyperdreambooth, wu2023tune, zhao2023motiondirector} that rely on text prompts that ``faithfully" describe the input image or video during model fine-tuning, our framework purposedly employs ``unfaithful" text prompts during the training phase.
Specifically, our approach involves the removal of background information.  
For instance, the text prompt `a cat is roaring on the grass under the tree' is simplified to `a cat is roaring' as presented in Fig.~\ref{fig-api}. This reduces background complexity as in Fig.~\ref{fig-api}a comapred to
Fig.~\ref{fig-api}b, facilitating the application of new appearance in
motion distillation.


\subsection{Inference Pipeline}
\label{sec: inference pipeline}
\quad Once  trained,
 in the inference phase, our process begins by computing inverted latents from the input video through DDIM inversion.
Subsequently, the inverted latents are fed into the temporally fine-tuned keyframe generation model, yielding 
short and low-resolution keyframes.
These keyframes then undergo temporal extension using the unaltered frame interpolation model.
Lastly, the interpolated frames are subjected to spatial enlargement through the spatial super-resolution model.
Overview of the process is depicted in Fig. \ref{fig-overview}.

\begin{figure*}[!htb]
    \centering
    \includegraphics[width=\textwidth]
    {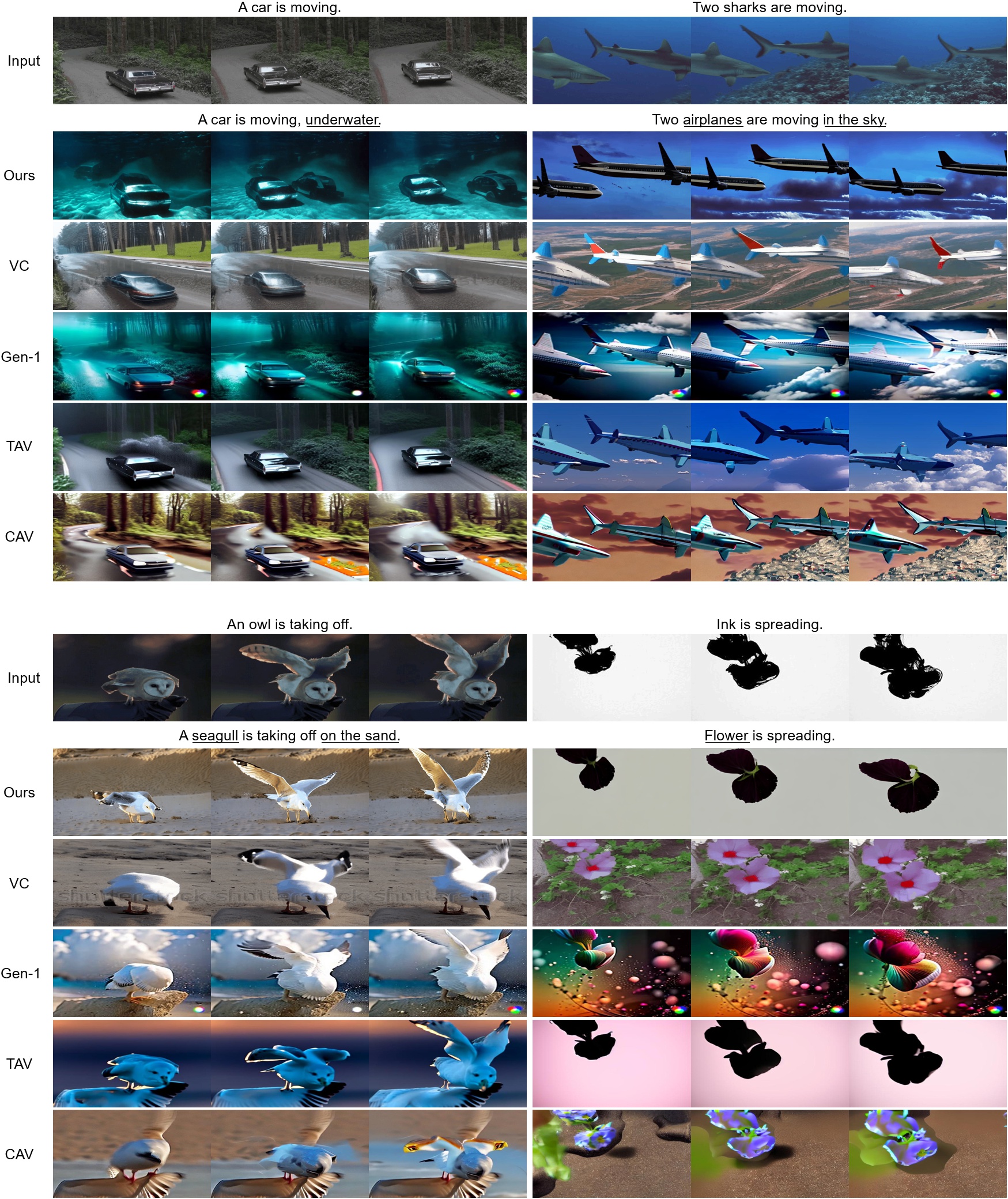}
    \caption{Qualitative comparison against state-of-the-art baselines. In contrast to other baselines, the proposed framework succeeds in motion-driven customization, even for difficult compositional customization.}
    \label{fig-qualitative}
\end{figure*}

\section{Experiments}
\subsection{Implementation Details}
\quad In our experiments, we choose Show-1 \cite{zhang2023show} as our VDM backbone and its publicly available pre-trained weights \footnote{\url{https://huggingface.co/showlab/show-1-base}}.
All experiments were conducted using a single NVIDIA RTX 6000 GPU.
VMC with Show-1 demonstrates efficient resource usage, requiring only 15GB of vRAM during mixed-precision training \cite{micikevicius2017mixed}, which is completed within 5 minutes.
During inference, generating a single video comprising 29 frames at a resolution of 576 x 320 consumes 18GB of vRAM and takes approximately 12 minutes.

\subsection{Baseline Comparisons}
\noindent \textbf{Dataset Selection.}
In our experiments, we draw upon a dataset that comprises 24 videos.
These videos encompass a broad spectrum of motion types occurring in various contexts, encompassing vehicles, humans, birds, plants, diffusion processes, mammals, sea creatures, and more.
This diversity provides a comprehensive range of motion scenarios for our assessment.
Out of these 24 videos, 13 are sourced from the DAVIS dataset \cite{pont20172017}, 10 from the WebVid dataset \cite{bain2021frozen}, and 1 video is obtained from LAMP \cite{wu2023lamp}.


\paragraph{Baselines.} 
Our method is compared against four contemporary baselines that integrate depth map signals into the diffusion denoising process to assimilate motion information. Notably, our approach operates without the necessity of depth maps during both training and inference, in contrast to these baseline methods.

Specifically, \textbf{VideoComposer} (VC) \cite{wang2023videocomposer} is an open-source latent-based video diffusion model tailored for compositional video generation tasks. \textbf{Gen-1} \citep{esser2023structure} introduces a video diffusion architecture incorporating additional structure and content guidance for video-to-video translation. In contrast to our targeted fine-tuning of temporal attention, \textbf{Tune-A-Video} (TAV) \cite{wu2023tune} fine-tunes self, cross, and temporal attention layers within a pre-trained, but inflated T2I model on input videos. \textbf{Control-A-Video} (CAV) \cite{chen2023control} introduces a controllable T2V diffusion model utilizing control signals and a first-frame conditioning strategy. Notably, while closely aligned with our framework, Motion Director \cite{zhao2023motiondirector} lacks available code at the time of our research.


\paragraph{Qualitative Results.}
We offer visual comparisons of our method against four baselines in Fig. \ref{fig-qualitative}.
The compared baselines face challenges in adapting the motion of the input video to new contexts.
They exhibit difficulties in applying the overall motion, neglecting the specific background indicated in the target text (e.g., ``underwater" or ``on the sand").
Additionally, they face difficulties in deviating from the original shape of the subject in the input video, leading to issues like a shark-shaped airplane, an owl-shaped seagull, or preservation of the shape of the ground where a seagull is taking off. In contrast, the proposed framework succeeds in motion-driven customization, even for difficult compositional customization, e.g. \texttt{Two sharks are moving.} $\rightarrow$ \texttt{Two \textit{airplanes} are moving \textit{in the sky}}.

\begin{figure*}[htb]
    \centering
    \vspace{-5pt}
    \includegraphics[width=\textwidth]
    {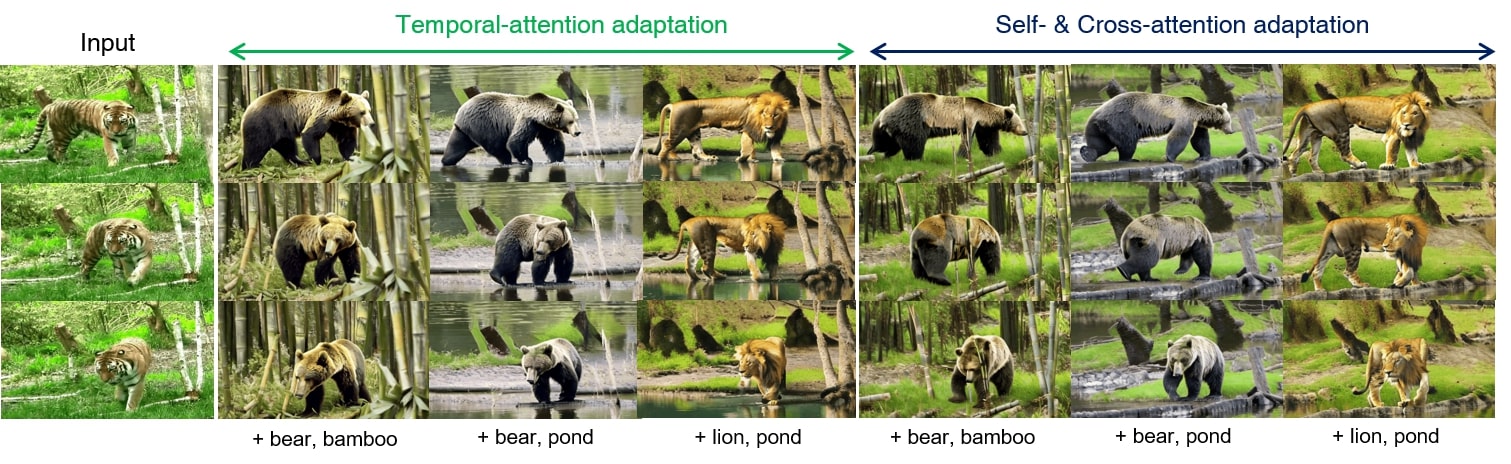}
    \caption{Comparative analysis of the proposed frameworks with fine-tuning (\textbf{a}) temporal attention and (\textbf{b}) self- and cross-attention layers.}
    \label{fig-ablation-attention}
\end{figure*}

\begin{figure*}[t!]
    \centering
    \vspace{-5pt}
    \includegraphics[width=\textwidth]
    {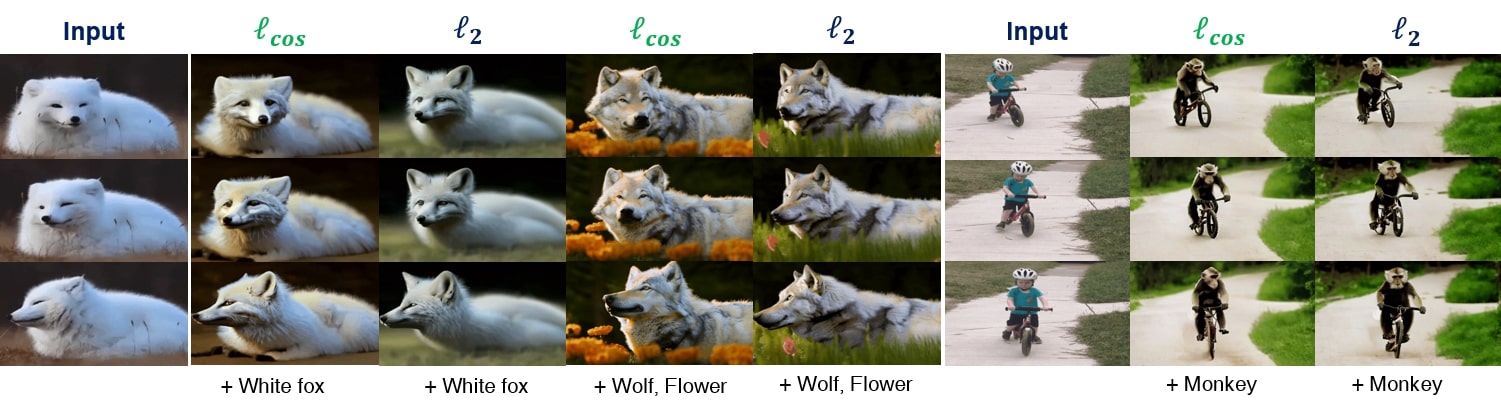}
    \caption{Comparative analysis of the proposed frameworks with (\textbf{a}) $\ell_{cos}$ and (\textbf{b}) $\ell_2$ loss functions.}
    \label{fig-ablation-mse}
\end{figure*}

\paragraph{Quantitative Results.} We further quantitatively demonstrate the effectiveness of our method against the baselines through automatic metrics and user study.

\noindent\textit{Automatic Metrics.} We use CLIP \citep{radford2021learning} encoders for automatic metrics.
For textual alignment, we compute the average cosine similarity between the target prompt and the generated frames.
In terms of frame consistency, we obtain CLIP image features within the output video and then calculate the average cosine similarity among all pairs of video frames.
For methods that generate temporally interpolated frames, we utilized the keyframe indexes to calculate the metric for a fair evaluation. 
To illustrate, in the case of VMC, which takes an 8-frame input and produces a 29-frame output, we considered the frames at the following indexes: \textit{1, 5, 9, 13, 17, 21, 25, 29}.
As shown in Table \ref{table-quantitative}, VMC outperforms baselines in both text alignment and temporal consistency.

\noindent\textit{User Study.} We conducted a survey involving a total of 27 participants to assess four key aspects: 
the preservation of motion between the input video and the generated output video, 
appearance diversity in the output video compared to the input video,
the text alignment with the target prompt, 
and the overall consistency of the generated frames.
The survey utilized a rating scale ranging from 1 to 5.
For assessing motion preservation, we employed the question: ``To what extent is the motion of the input video retained in the output video?"
To evaluate appearance diversity, participants were asked: ``To what extent does the appearance of the output video avoid being restricted on the input video's appearance?"
Tab. \ref{table-quantitative} shows that our method surpasses the baselines in all four aspects.

\begin{table}[h]
\centering
\vspace{-5pt}
\resizebox{\columnwidth}{!}{
\begin{tabular}{@{\extracolsep{0pt}} c|cc|cc ccc@{}}
\hline
        & Text      & Temporal    & Motion       & Appearance  & Text      & Temporal \\
        & Alignment & Consistency & Preservation & Diversity  & Alignment & Consistency \\
\hline
VC      & 0.798 & 0.958 & 3.45 & 3.43 & 2.96 & 3.03\\
Gen-1   & 0.780 & 0.957 & 3.46 & 3.17 & 2.87 & 2.73\\
TAV     & 0.758 & 0.947 & 3.50 & 2.88 & 2.67 & 2.80\\
CAV     & 0.764 & 0.952 & 2.75 & 2.45 & 2.07 & 2.00\\
Ours    & \textbf{0.801} & \textbf{0.959} & \textbf{4.42} & \textbf{4.54} & \textbf{4.56} & \textbf{4.57} \\
\hline
\end{tabular}
}
\caption{Quantitative evaluation using CLIP and user study. Our method significantly outperforms the other baselines.}
\label{table-quantitative}
\end{table}

\begin{figure*}[htb]
    \centering
    \vspace{-5pt}
    \includegraphics[width=\textwidth]
    {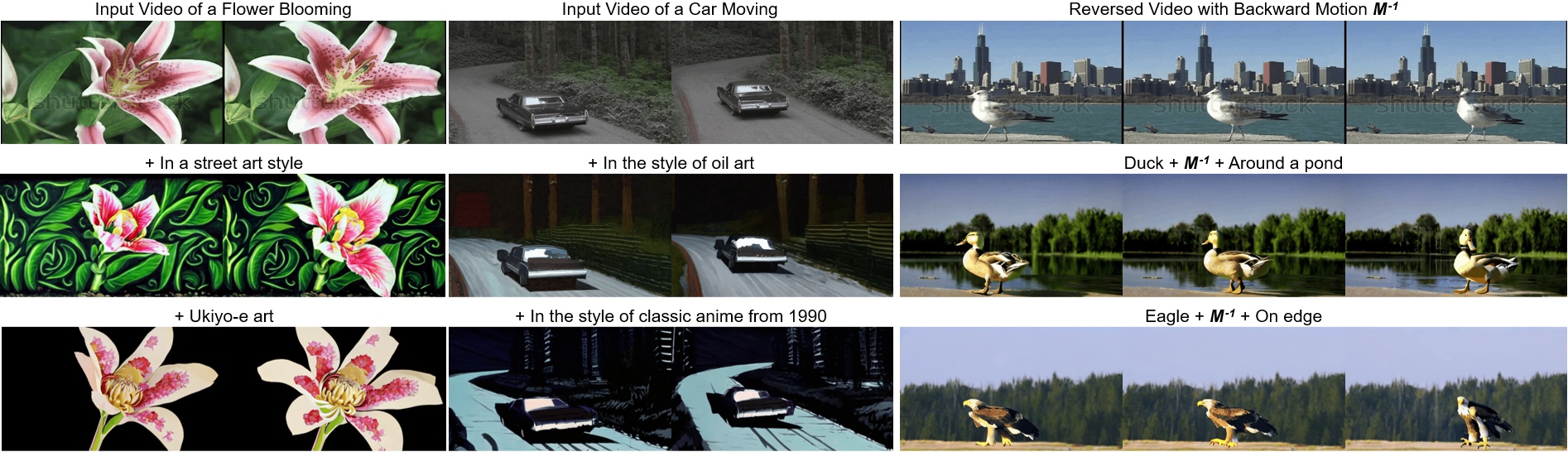}
    \caption{\textit{Left}: Style transfer on two videos. \textit{Right}: Motion customization results on the video of ``A seagull is walking \textit{backward}."}
    \label{fig-style-backward}
\end{figure*}

\subsection{Ablation Studies}
\label{sec: ablation}
\noindent\textbf{Comparisons on attention layers.} 
We conducted a comparative study evaluating the performance of fine-tuning: (\textbf{a}) temporal attention layers and (\textbf{b}) self- and cross-attention layers. Illustrated in Fig. \ref{fig-ablation-attention}, both frameworks exhibit proficient motion learning capabilities. Notably, the utilization of customized temporal attention layers (\textbf{a}) yields smoother frame transitions, indicating the effectiveness of the optimization framework \eqref{eq: optimization framework} in encouraging motion distillation, with a slight preference observed for customized temporal attention layers.

This observation stems from the premise that integrating the proposed motion distillation objective \eqref{eq: optimization framework} may autonomously and accurately embed motion information within temporal attention layers \cite{ho2204video, ho2022imagen}. This suggests a potential application of the motion distillation objective for training large-scale video diffusion models, warranting further exploration in future research endeavors.

\vspace{-5pt}
\paragraph{Choice of loss functions.} 
In addition, we conducted a comparative analysis on distinct training loss functions in \eqref{eq: optimization framework}: the $\ell_2$-distance and $\ell_{cos}$ as delineated in \eqref{eq: optimization framework}. As depicted in Fig. \ref{fig-ablation-mse}, the $\delta \epsilonb$-matching process in \eqref{eq: v0 align} and \eqref{eq: optimization framework} demonstrates compatibility with generic loss functions. While both $\ell_2(\delta \epsilonb_t^n, \delta \epsilonb_{\theta, t}^n)$ and $\ell_{cos}(\delta \epsilonb_t^n, \delta \epsilonb_{\theta, t}^n)$ are promising objectives, the marginal superiority of $\ell_{cos}(\delta \epsilonb_t^n, \delta \epsilonb_{\theta, t}^n)$ led to its adoption for visualizations in this study.

\begin{figure}
    \centering
    \vspace{-5pt}
    \includegraphics[width=\linewidth]{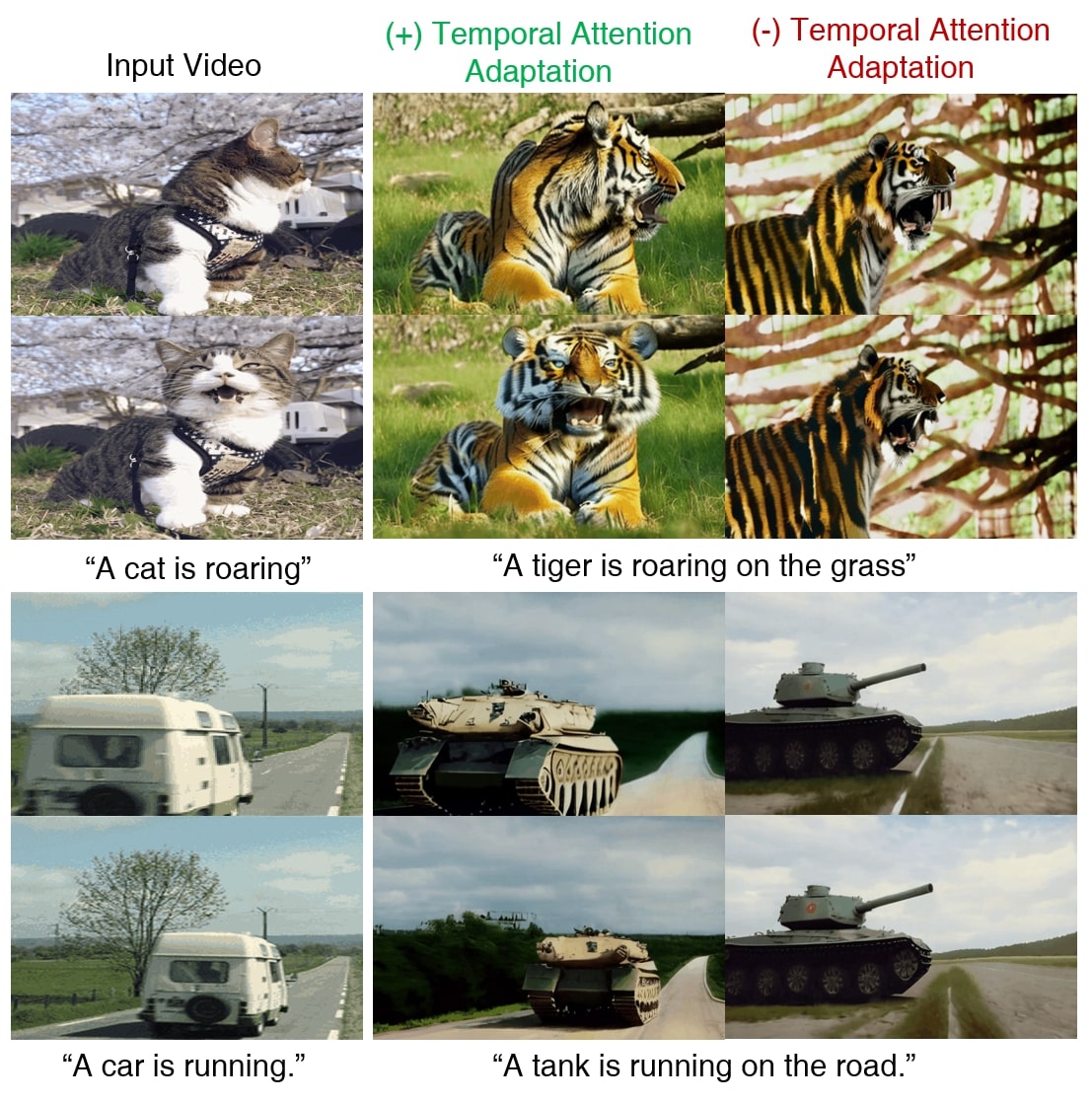}
    \caption{Ablation study on temporal attention adaptation. Without temporal attention adaptation, motion distillation fails.}
    \label{fig-ablation-adaptation}
\end{figure}

\paragraph{Importance of adaptation.} 
To assess the importance of temporal attention adaptation, we conducted a visualization of customized generations without temporal attention adaptation, as detailed in Section \ref{sec: temporal adaptation}. Specifically, from our original architecture in Fig. \ref{fig-overview}, we omitted attention adaptation and performed inference by maintaining the U-Net modules in a frozen state. The outcomes depicted in Fig. \ref{fig-ablation-adaptation} indicate that while DDIM inversion guides the generations to mimic the motion of the input video, it alone does not ensure successful motion distillation. The observed changes in appearance and motion exhibit an entangled relationship. Consequently, this underlines the necessity of an explicit motion distillation objective to achieve consistent motion transfer, independent of any alterations in appearance.

\subsection{Additional results}
\noindent\textbf{Video Style Transfer.} We illustrate video style transfer applications in Fig.~\ref{fig-style-backward}-\textit{Left}.
We incorporate style prompts at the end of the text after applying appearance-invariant prompting (see Section~\ref{method-prompting}).
Target styles are fluidly injected while preserving the distilled motion of an input video.

\noindent\textbf{Learning Backward Motion.} To further verify our video motion customization capabilities, we present a challenging scenario: extracting backward motion from a reversed video sequence where frames are arranged in reverse order. This scenario, an exceedingly rare event in real-world videos, is highly improbable within standard training video datasets \cite{bain2021frozen}. Illustrated in Fig. \ref{fig-style-backward}, our VMC framework showcases proficiency in learning ``a bird walking backward" motion and generating diverse videos with distinct subjects and backgrounds. This capability not only enables leveraging the distilled motion but also offers prospects for further contextual editing.

\section{Conclusion}
\vspace{-5pt}
\quad This paper introduces Video Motion Customization (VMC), addressing challenges in adapting Text-to-Video (T2V) models to generate motion-driven diverse visual customizations. 
Existing models struggle with accurately replicating motion from a target video and creating varied visual outputs, leading to entanglements of appearance and motion data. 
To overcome this, our VMC framework presents a novel one-shot tuning approach, focusing on adapting temporal attention layers within video diffusion models. 
This framework stands out for its efficiency in time and memory, ease of implementation, and minimal hyperparameters. We demonstrated the efficacy of our customization methods across diverse motion types, appearances, and contexts. 

\noindent\textbf{Ethics Statement.}
Our work is based on a generative model with potential for misuse, including the creation of deceptive content, which may have negative societal impacts.
Additionally, inappropriate content and biases could be included in the datasets used for the foundational training.

{
    \small
    \bibliographystyle{ieeenat_fullname}
    \bibliography{main}

\begin{thebibliography}{39}
\providecommand{\natexlab}[1]{#1}
\providecommand{\url}[1]{\texttt{#1}}
\expandafter\ifx\csname urlstyle\endcsname\relax
  \providecommand{\doi}[1]{doi: #1}\else
  \providecommand{\doi}{doi: \begingroup \urlstyle{rm}\Url}\fi

\bibitem[Bain et~al.(2021)Bain, Nagrani, Varol, and Zisserman]{bain2021frozen}
Max Bain, Arsha Nagrani, G{\"u}l Varol, and Andrew Zisserman.
\newblock Frozen in time: A joint video and image encoder for end-to-end retrieval.
\newblock In \emph{Proceedings of the IEEE/CVF International Conference on Computer Vision}, pages 1728--1738, 2021.

\bibitem[Blattmann et~al.(2023)Blattmann, Rombach, Ling, Dockhorn, Kim, Fidler, and Kreis]{blattmann2023align}
Andreas Blattmann, Robin Rombach, Huan Ling, Tim Dockhorn, Seung~Wook Kim, Sanja Fidler, and Karsten Kreis.
\newblock Align your latents: High-resolution video synthesis with latent diffusion models.
\newblock In \emph{Proceedings of the IEEE/CVF Conference on Computer Vision and Pattern Recognition}, pages 22563--22575, 2023.

\bibitem[Byeon et~al.(2022)Byeon, Park, Kim, Lee, Baek, and Kim]{kakaobrain2022coyo-700m}
Minwoo Byeon, Beomhee Park, Haecheon Kim, Sungjun Lee, Woonhyuk Baek, and Saehoon Kim.
\newblock Coyo-700m: Image-text pair dataset.
\newblock \url{https://github.com/kakaobrain/coyo-dataset}, 2022.

\bibitem[Chen et~al.(2023{\natexlab{a}})Chen, Wang, Zeng, Zhang, Zhou, Han, and Zhu]{chen2023videodreamer}
Hong Chen, Xin Wang, Guanning Zeng, Yipeng Zhang, Yuwei Zhou, Feilin Han, and Wenwu Zhu.
\newblock Videodreamer: Customized multi-subject text-to-video generation with disen-mix finetuning.
\newblock \emph{arXiv preprint arXiv:2311.00990}, 2023{\natexlab{a}}.

\bibitem[Chen et~al.(2023{\natexlab{b}})Chen, Wu, Xie, Wu, Li, Xia, Xiao, and Lin]{chen2023control}
Weifeng Chen, Jie Wu, Pan Xie, Hefeng Wu, Jiashi Li, Xin Xia, Xuefeng Xiao, and Liang Lin.
\newblock Control-a-video: Controllable text-to-video generation with diffusion models.
\newblock \emph{arXiv preprint arXiv:2305.13840}, 2023{\natexlab{b}}.

\bibitem[Efron(2011)]{efron2011tweedie}
Bradley Efron.
\newblock Tweedie’s formula and selection bias.
\newblock \emph{Journal of the American Statistical Association}, 106\penalty0 (496):\penalty0 1602--1614, 2011.

\bibitem[Esser et~al.(2023)Esser, Chiu, Atighehchian, Granskog, and Germanidis]{esser2023structure}
Patrick Esser, Johnathan Chiu, Parmida Atighehchian, Jonathan Granskog, and Anastasis Germanidis.
\newblock Structure and content-guided video synthesis with diffusion models.
\newblock In \emph{Proceedings of the IEEE/CVF International Conference on Computer Vision}, pages 7346--7356, 2023.

\bibitem[Gal et~al.(2022)Gal, Alaluf, Atzmon, Patashnik, Bermano, Chechik, and Cohen-Or]{gal2022image}
Rinon Gal, Yuval Alaluf, Yuval Atzmon, Or Patashnik, Amit~H Bermano, Gal Chechik, and Daniel Cohen-Or.
\newblock An image is worth one word: Personalizing text-to-image generation using textual inversion.
\newblock \emph{arXiv preprint arXiv:2208.01618}, 2022.

\bibitem[Ge et~al.(2023)Ge, Nah, Liu, Poon, Tao, Catanzaro, Jacobs, Huang, Liu, and Balaji]{ge2023preserve}
Songwei Ge, Seungjun Nah, Guilin Liu, Tyler Poon, Andrew Tao, Bryan Catanzaro, David Jacobs, Jia-Bin Huang, Ming-Yu Liu, and Yogesh Balaji.
\newblock Preserve your own correlation: A noise prior for video diffusion models.
\newblock In \emph{Proceedings of the IEEE/CVF International Conference on Computer Vision}, pages 22930--22941, 2023.

\bibitem[Gu et~al.(2023)Gu, Wang, Wu, Shi, Chen, Fan, Xiao, Zhao, Chang, Wu, et~al.]{gu2023mix}
Yuchao Gu, Xintao Wang, Jay~Zhangjie Wu, Yujun Shi, Yunpeng Chen, Zihan Fan, Wuyou Xiao, Rui Zhao, Shuning Chang, Weijia Wu, et~al.
\newblock Mix-of-show: Decentralized low-rank adaptation for multi-concept customization of diffusion models.
\newblock \emph{arXiv preprint arXiv:2305.18292}, 2023.

\bibitem[Han et~al.(2023)Han, Li, Zhang, Milanfar, Metaxas, and Yang]{han2023svdiff}
Ligong Han, Yinxiao Li, Han Zhang, Peyman Milanfar, Dimitris Metaxas, and Feng Yang.
\newblock Svdiff: Compact parameter space for diffusion fine-tuning.
\newblock \emph{arXiv preprint arXiv:2303.11305}, 2023.

\bibitem[Ho et~al.()Ho, Salimans, Gritsenko, Chan, Norouzi, and Fleet]{ho2204video}
J Ho, T Salimans, A Gritsenko, W Chan, M Norouzi, and DJ Fleet.
\newblock Video diffusion models. arxiv 2022.
\newblock \emph{arXiv preprint arXiv:2204.03458}.

\bibitem[Ho et~al.(2020)Ho, Jain, and Abbeel]{ho2020denoising}
Jonathan Ho, Ajay Jain, and Pieter Abbeel.
\newblock Denoising diffusion probabilistic models.
\newblock \emph{Advances in neural information processing systems}, 33:\penalty0 6840--6851, 2020.

\bibitem[Ho et~al.(2022)Ho, Chan, Saharia, Whang, Gao, Gritsenko, Kingma, Poole, Norouzi, Fleet, et~al.]{ho2022imagen}
Jonathan Ho, William Chan, Chitwan Saharia, Jay Whang, Ruiqi Gao, Alexey Gritsenko, Diederik~P Kingma, Ben Poole, Mohammad Norouzi, David~J Fleet, et~al.
\newblock Imagen video: High definition video generation with diffusion models.
\newblock \emph{arXiv preprint arXiv:2210.02303}, 2022.

\bibitem[Hu et~al.(2021)Hu, Shen, Wallis, Allen-Zhu, Li, Wang, Wang, and Chen]{hu2021lora}
Edward~J Hu, Yelong Shen, Phillip Wallis, Zeyuan Allen-Zhu, Yuanzhi Li, Shean Wang, Lu Wang, and Weizhu Chen.
\newblock Lora: Low-rank adaptation of large language models.
\newblock \emph{arXiv preprint arXiv:2106.09685}, 2021.

\bibitem[Hyv{\"a}rinen and Dayan(2005)]{hyvarinen2005estimation}
Aapo Hyv{\"a}rinen and Peter Dayan.
\newblock Estimation of non-normalized statistical models by score matching.
\newblock \emph{Journal of Machine Learning Research}, 6\penalty0 (4), 2005.

\bibitem[Loshchilov and Hutter(2017)]{loshchilov2017decoupled}
Ilya Loshchilov and Frank Hutter.
\newblock Decoupled weight decay regularization.
\newblock \emph{arXiv preprint arXiv:1711.05101}, 2017.

\bibitem[Lu et~al.(2023)Lu, Tunanyan, Wang, Navasardyan, Wang, and Shi]{lu2023specialist}
Haoming Lu, Hazarapet Tunanyan, Kai Wang, Shant Navasardyan, Zhangyang Wang, and Humphrey Shi.
\newblock Specialist diffusion: Plug-and-play sample-efficient fine-tuning of text-to-image diffusion models to learn any unseen style.
\newblock In \emph{Proceedings of the IEEE/CVF Conference on Computer Vision and Pattern Recognition}, pages 14267--14276, 2023.

\bibitem[Ma et~al.(2023)Ma, Liang, Chen, and Lu]{ma2023subject}
Jian Ma, Junhao Liang, Chen Chen, and Haonan Lu.
\newblock Subject-diffusion: Open domain personalized text-to-image generation without test-time fine-tuning.
\newblock \emph{arXiv preprint arXiv:2307.11410}, 2023.

\bibitem[Micikevicius et~al.(2017)Micikevicius, Narang, Alben, Diamos, Elsen, Garcia, Ginsburg, Houston, Kuchaiev, Venkatesh, et~al.]{micikevicius2017mixed}
Paulius Micikevicius, Sharan Narang, Jonah Alben, Gregory Diamos, Erich Elsen, David Garcia, Boris Ginsburg, Michael Houston, Oleksii Kuchaiev, Ganesh Venkatesh, et~al.
\newblock Mixed precision training.
\newblock \emph{arXiv preprint arXiv:1710.03740}, 2017.

\bibitem[Pont-Tuset et~al.(2017)Pont-Tuset, Perazzi, Caelles, Arbel{\'a}ez, Sorkine-Hornung, and Van~Gool]{pont20172017}
Jordi Pont-Tuset, Federico Perazzi, Sergi Caelles, Pablo Arbel{\'a}ez, Alex Sorkine-Hornung, and Luc Van~Gool.
\newblock The 2017 davis challenge on video object segmentation.
\newblock \emph{arXiv preprint arXiv:1704.00675}, 2017.

\bibitem[Radford et~al.(2021)Radford, Kim, Hallacy, Ramesh, Goh, Agarwal, Sastry, Askell, Mishkin, Clark, et~al.]{radford2021learning}
Alec Radford, Jong~Wook Kim, Chris Hallacy, Aditya Ramesh, Gabriel Goh, Sandhini Agarwal, Girish Sastry, Amanda Askell, Pamela Mishkin, Jack Clark, et~al.
\newblock Learning transferable visual models from natural language supervision.
\newblock In \emph{International conference on machine learning}, pages 8748--8763. PMLR, 2021.

\bibitem[Ruiz et~al.(2023{\natexlab{a}})Ruiz, Li, Jampani, Pritch, Rubinstein, and Aberman]{ruiz2023dreambooth}
Nataniel Ruiz, Yuanzhen Li, Varun Jampani, Yael Pritch, Michael Rubinstein, and Kfir Aberman.
\newblock Dreambooth: Fine tuning text-to-image diffusion models for subject-driven generation.
\newblock In \emph{Proceedings of the IEEE/CVF Conference on Computer Vision and Pattern Recognition}, pages 22500--22510, 2023{\natexlab{a}}.

\bibitem[Ruiz et~al.(2023{\natexlab{b}})Ruiz, Li, Jampani, Wei, Hou, Pritch, Wadhwa, Rubinstein, and Aberman]{ruiz2023hyperdreambooth}
Nataniel Ruiz, Yuanzhen Li, Varun Jampani, Wei Wei, Tingbo Hou, Yael Pritch, Neal Wadhwa, Michael Rubinstein, and Kfir Aberman.
\newblock Hyperdreambooth: Hypernetworks for fast personalization of text-to-image models.
\newblock \emph{arXiv preprint arXiv:2307.06949}, 2023{\natexlab{b}}.

\bibitem[Schuhmann et~al.(2022)Schuhmann, Beaumont, Vencu, Gordon, Wightman, Cherti, Coombes, Katta, Mullis, Wortsman, et~al.]{schuhmann2022laion}
Christoph Schuhmann, Romain Beaumont, Richard Vencu, Cade Gordon, Ross Wightman, Mehdi Cherti, Theo Coombes, Aarush Katta, Clayton Mullis, Mitchell Wortsman, et~al.
\newblock Laion-5b: An open large-scale dataset for training next generation image-text models.
\newblock \emph{Advances in Neural Information Processing Systems}, 35:\penalty0 25278--25294, 2022.

\bibitem[Shi et~al.(2023)Shi, Xiong, Lin, and Jung]{shi2023instantbooth}
Jing Shi, Wei Xiong, Zhe Lin, and Hyun~Joon Jung.
\newblock Instantbooth: Personalized text-to-image generation without test-time finetuning.
\newblock \emph{arXiv preprint arXiv:2304.03411}, 2023.

\bibitem[Singer et~al.(2022)Singer, Polyak, Hayes, Yin, An, Zhang, Hu, Yang, Ashual, Gafni, et~al.]{singer2022make}
Uriel Singer, Adam Polyak, Thomas Hayes, Xi Yin, Jie An, Songyang Zhang, Qiyuan Hu, Harry Yang, Oron Ashual, Oran Gafni, et~al.
\newblock Make-a-video: Text-to-video generation without text-video data.
\newblock \emph{arXiv preprint arXiv:2209.14792}, 2022.

\bibitem[Sohl-Dickstein et~al.(2015)Sohl-Dickstein, Weiss, Maheswaranathan, and Ganguli]{sohl2015deep}
Jascha Sohl-Dickstein, Eric Weiss, Niru Maheswaranathan, and Surya Ganguli.
\newblock Deep unsupervised learning using nonequilibrium thermodynamics.
\newblock In \emph{International conference on machine learning}, pages 2256--2265. PMLR, 2015.

\bibitem[Song et~al.(2020{\natexlab{a}})Song, Meng, and Ermon]{song2020denoising}
Jiaming Song, Chenlin Meng, and Stefano Ermon.
\newblock Denoising diffusion implicit models.
\newblock \emph{arXiv preprint arXiv:2010.02502}, 2020{\natexlab{a}}.

\bibitem[Song and Ermon(2019)]{song2019generative}
Yang Song and Stefano Ermon.
\newblock Generative modeling by estimating gradients of the data distribution.
\newblock \emph{Advances in neural information processing systems}, 32, 2019.

\bibitem[Song et~al.(2020{\natexlab{b}})Song, Sohl-Dickstein, Kingma, Kumar, Ermon, and Poole]{song2020score}
Yang Song, Jascha Sohl-Dickstein, Diederik~P Kingma, Abhishek Kumar, Stefano Ermon, and Ben Poole.
\newblock Score-based generative modeling through stochastic differential equations.
\newblock \emph{arXiv preprint arXiv:2011.13456}, 2020{\natexlab{b}}.

\bibitem[Wang et~al.(2023{\natexlab{a}})Wang, Yuan, Zhang, Chen, Wang, Zhang, Shen, Zhao, and Zhou]{wang2023videocomposer}
Xiang Wang, Hangjie Yuan, Shiwei Zhang, Dayou Chen, Jiuniu Wang, Yingya Zhang, Yujun Shen, Deli Zhao, and Jingren Zhou.
\newblock Videocomposer: Compositional video synthesis with motion controllability.
\newblock \emph{arXiv preprint arXiv:2306.02018}, 2023{\natexlab{a}}.

\bibitem[Wang et~al.(2023{\natexlab{b}})Wang, Chen, Ma, Zhou, Huang, Wang, Yang, He, Yu, Yang, et~al.]{wang2023lavie}
Yaohui Wang, Xinyuan Chen, Xin Ma, Shangchen Zhou, Ziqi Huang, Yi Wang, Ceyuan Yang, Yinan He, Jiashuo Yu, Peiqing Yang, et~al.
\newblock Lavie: High-quality video generation with cascaded latent diffusion models.
\newblock \emph{arXiv preprint arXiv:2309.15103}, 2023{\natexlab{b}}.

\bibitem[Wei et~al.(2023)Wei, Zhang, Ji, Bai, Zhang, and Zuo]{wei2023elite}
Yuxiang Wei, Yabo Zhang, Zhilong Ji, Jinfeng Bai, Lei Zhang, and Wangmeng Zuo.
\newblock Elite: Encoding visual concepts into textual embeddings for customized text-to-image generation.
\newblock \emph{arXiv preprint arXiv:2302.13848}, 2023.

\bibitem[Wu et~al.(2023{\natexlab{a}})Wu, Ge, Wang, Lei, Gu, Shi, Hsu, Shan, Qie, and Shou]{wu2023tune}
Jay~Zhangjie Wu, Yixiao Ge, Xintao Wang, Stan~Weixian Lei, Yuchao Gu, Yufei Shi, Wynne Hsu, Ying Shan, Xiaohu Qie, and Mike~Zheng Shou.
\newblock Tune-a-video: One-shot tuning of image diffusion models for text-to-video generation.
\newblock In \emph{Proceedings of the IEEE/CVF International Conference on Computer Vision}, pages 7623--7633, 2023{\natexlab{a}}.

\bibitem[Wu et~al.(2023{\natexlab{b}})Wu, Chen, Yang, Guo, Li, and Zhang]{wu2023lamp}
Ruiqi Wu, Liangyu Chen, Tong Yang, Chunle Guo, Chongyi Li, and Xiangyu Zhang.
\newblock Lamp: Learn a motion pattern for few-shot-based video generation.
\newblock \emph{arXiv preprint arXiv:2310.10769}, 2023{\natexlab{b}}.

\bibitem[Zhang et~al.(2023)Zhang, Wu, Liu, Zhao, Ran, Gu, Gao, and Shou]{zhang2023show}
David~Junhao Zhang, Jay~Zhangjie Wu, Jia-Wei Liu, Rui Zhao, Lingmin Ran, Yuchao Gu, Difei Gao, and Mike~Zheng Shou.
\newblock Show-1: Marrying pixel and latent diffusion models for text-to-video generation.
\newblock \emph{arXiv preprint arXiv:2309.15818}, 2023.

\bibitem[Zhao et~al.(2023)Zhao, Gu, Wu, Zhang, Liu, Wu, Keppo, and Shou]{zhao2023motiondirector}
Rui Zhao, Yuchao Gu, Jay~Zhangjie Wu, David~Junhao Zhang, Jiawei Liu, Weijia Wu, Jussi Keppo, and Mike~Zheng Shou.
\newblock Motiondirector: Motion customization of text-to-video diffusion models.
\newblock \emph{arXiv preprint arXiv:2310.08465}, 2023.

\bibitem[Zhou et~al.(2022)Zhou, Wang, Yan, Lv, Zhu, and Feng]{zhou2022magicvideo}
Daquan Zhou, Weimin Wang, Hanshu Yan, Weiwei Lv, Yizhe Zhu, and Jiashi Feng.
\newblock Magicvideo: Efficient video generation with latent diffusion models.
\newblock \emph{arXiv preprint arXiv:2211.11018}, 2022.

\end{thebibliography}
}

\clearpage

\appendix
\section{Appendix}
The supplementary sections in Appendix are organized as follows. 
Section \ref{sec: pseudocode} introduces the pseudo training algorithm behind our Video Motion Customization (VMC) framework.
In Section \ref{sec: related works}, we provide a discussion on related works in the field of generative model customization.
Following this, we delve into the details on our training and inference configurations in Section \ref{sec: training details}. 
Concluding the document, Section \ref{sec: additional results} features a showcase of additional results obtained from our VMC framework.

\section{Pseudo Training Algorithm}
\label{sec: pseudocode}
\begin{algorithm}
\label{pseudocode}
	\caption{Temporal Attention Adaption} 
	\begin{algorithmic}[1]
	    \State {\bfseries Input:} $N$-frame input video sequence $(\vb_0^n)_{n \in \{1, \dots, N\}}$, appearance-invariant training prompt $\Pc_{\text{inv}}$, textual encoder $\psi$, Training iterations $M$, key-frame generator parameterized by $\theta$ and its temporal attention parameters $\theta_{\text{TA}}$. 
     
        \State {\bfseries Output:} Fine-tuned temporal attention layers \(\theta_{\text{TA}}^*\). \\

            \For {$step=1$ \textbf{to} $M$}
                \State Sample timestep $t \in [0, T]$ and Gaussian noise $\epsilonb_t^{1:N}$, where \(\epsilonb_t^n \in \mathbb{R}^d \sim \Nc(0, I)\)
                \State Prepare text embeddings $c_{\text{inv}} = \psi(\Pc_{\text{inv}})$
                \State $\vb_t^n = \sqrt{\bar{\alpha}_t} \vb_0^n + \sqrt{1 - \bar{\alpha}_t} \epsilonb_t^{n}$, $\forall n$.
                \State $\delta \epsilonb_{\theta, t}^n = \epsilonb_\theta^{n+1}(\vb_t^{1:N}, t, c_{\text{inv}}) - \epsilonb_\theta^n (\vb_t^{1:N}, t, c_{\text{inv}})$, $\forall n \leq N-1$. 
                \State $\delta \epsilonb_{t}^n = \epsilonb_t^{n+1} - \epsilonb_t^n$, $\forall n \leq N-1$ 
                \State Update $\theta_{\text{TA}}$ with $\frac{1}{N-1} \sum_{n} \ell_{\text{cos}} (\delta \epsilonb_t^n, \delta \epsilonb_{\theta, t}^n)$ 
            \EndFor
            
	\end{algorithmic} 
\end{algorithm}

We express Gaussian noises as $\epsilonb_t^{1:N}$ to avoid confusion. In our observation, aligning the image-space residuals ($\delta \vb_0^n$ and $\delta \hat{\vb}_0^n(t)$) corresponds to aligning the latent-space epsilon residuals ($\delta \epsilonb_t^n$ and $\delta \epsilonb_{\theta, t}^n$) across varying time steps $t \in [0, T]$. This relationship stems from expressing the motion vector $\delta \vb_0^n$ and its estimation $\delta \hat{\vb}_0^n(t)$ in terms of $\delta \vb_t^n$, $\delta \epsilonb_t^n$, and $\delta \epsilonb_{\theta, t}^n$. Consequently, the proposed optimization framework fine-tunes temporal attention layers by leveraging diverse diffusion latent spaces at time $t$ which potentially contains multi-scale rich descriptions of video frames. Therefore, this optimization approach seamlessly applies to video diffusion models trained using epsilon-matching, thanks to the equivalence between $\delta \epsilonb_t^n$-matching and $\delta \vb_0^n$-matching. 

\section{Related Works}
\label{sec: related works}
\quad\textbf{Image Customization.}
Prior methodologies in text-to-image customization, termed personalization \cite{gal2022image, ruiz2023dreambooth, shi2023instantbooth, ruiz2023hyperdreambooth, han2023svdiff, lu2023specialist, wei2023elite, gu2023mix}, aimed at capturing specific subject appearances while maintaining the model's ability to generate varied contents. 
However, this pursuit of personalization poses challenges in time and memory demands \cite{ruiz2023dreambooth}. 
Fine-tuning each personalized model requires substantial time costs while storing multiple personalized models may strain storage capacity. 
To address these hurdles, some approaches prioritize efficient parameter customization, leveraging techniques like LoRA \cite{hu2021lora, gu2023mix} or HyperNetwork \cite{ruiz2023hyperdreambooth} rather than training the entire model.

\textbf{Video Customization.} Building on the success of text-to-image customization, recent efforts have adopted text-to-image or text-to-video diffusion models for customizing videos in terms of appearance or motion. These endeavors, such as frameworks proposed by \cite{wu2023tune, chen2023videodreamer}, focus on creating videos faithful to given subjects or motions. 
Moreover, works by \cite{zhao2023motiondirector} or \cite{wu2023lamp} delve into motion-centric video customization, employing various fine-tuning approaches ranging from temporal-spatial motion learning layers to newly introduced LoRAs. 
In this paper, the proposed VMC framework emphasizes efficient motion customization with explicit motion distillation objectives, specifically targeting temporal attention layers. This approach, facilitated by cascaded video diffusion models, efficiently distills motion from a single video clip while minimizing computational burdens in terms of both time and memory.

\section{Training \& Inference Details}
\label{sec: training details}
\quad For our work, we utilize the cascaded video diffusion models from Show-1 \cite{zhang2023show}, employing its publicly accessible pre-trained weights \footnote{\url{https://huggingface.co/showlab}}.
Our approach maintains the temporal interpolation and spatial super-resolution modules in their original state while focusing our temporal optimization solely on the keyframe generator.
In specific, we finetune Query, Key, Value projection matrices $W^Q, W^K, W^V$ of temporal attention layers of the keyframe UNet.
We use AdamW \cite{loshchilov2017decoupled} optimizer, with weight decay of $0.01$ and learning rate $0.0001$.
By default, we employ 400 training steps.
During the inference phase, we perform DDIM inversion \cite{song2020denoising} for $75$ steps.
For the temporal interpolation and spatial super resolution stages, we follow the default settings of Show-1.

\section{Additional Results}
\begin{figure*}[htbp]
\centering
\includegraphics[width=0.9\textwidth]{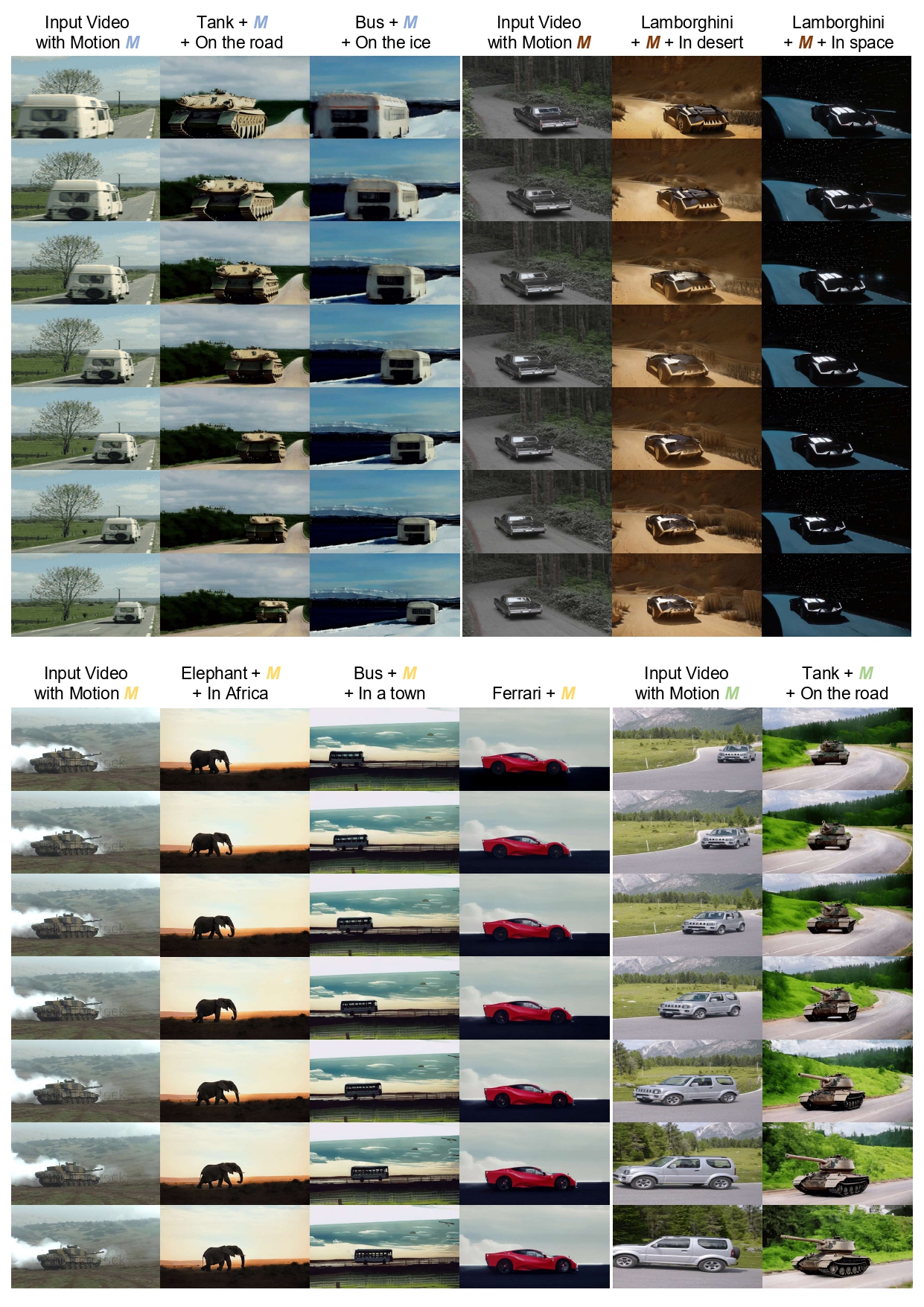}
\caption{Video Motion Customization results: Keyframes visualized.}
\label{fig:part1}
\end{figure*}

\begin{figure*}[htbp]
\centering
\includegraphics[width=0.9\textwidth]{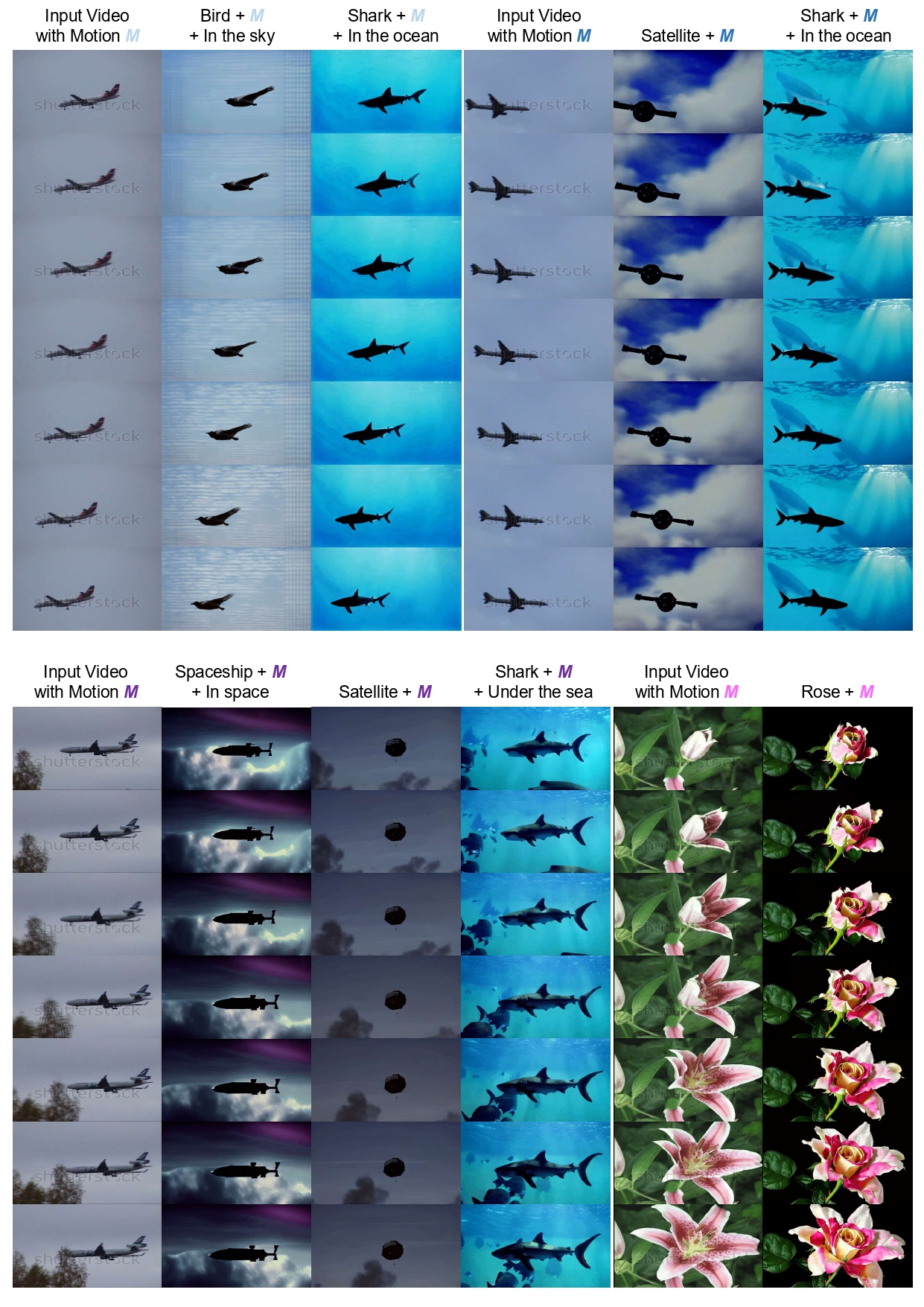}
\caption{Video Motion Customization results: Keyframes visualized.}
\label{fig:part2}
\end{figure*}

\begin{figure*}[htbp]
\centering
\includegraphics[width=0.9\textwidth]{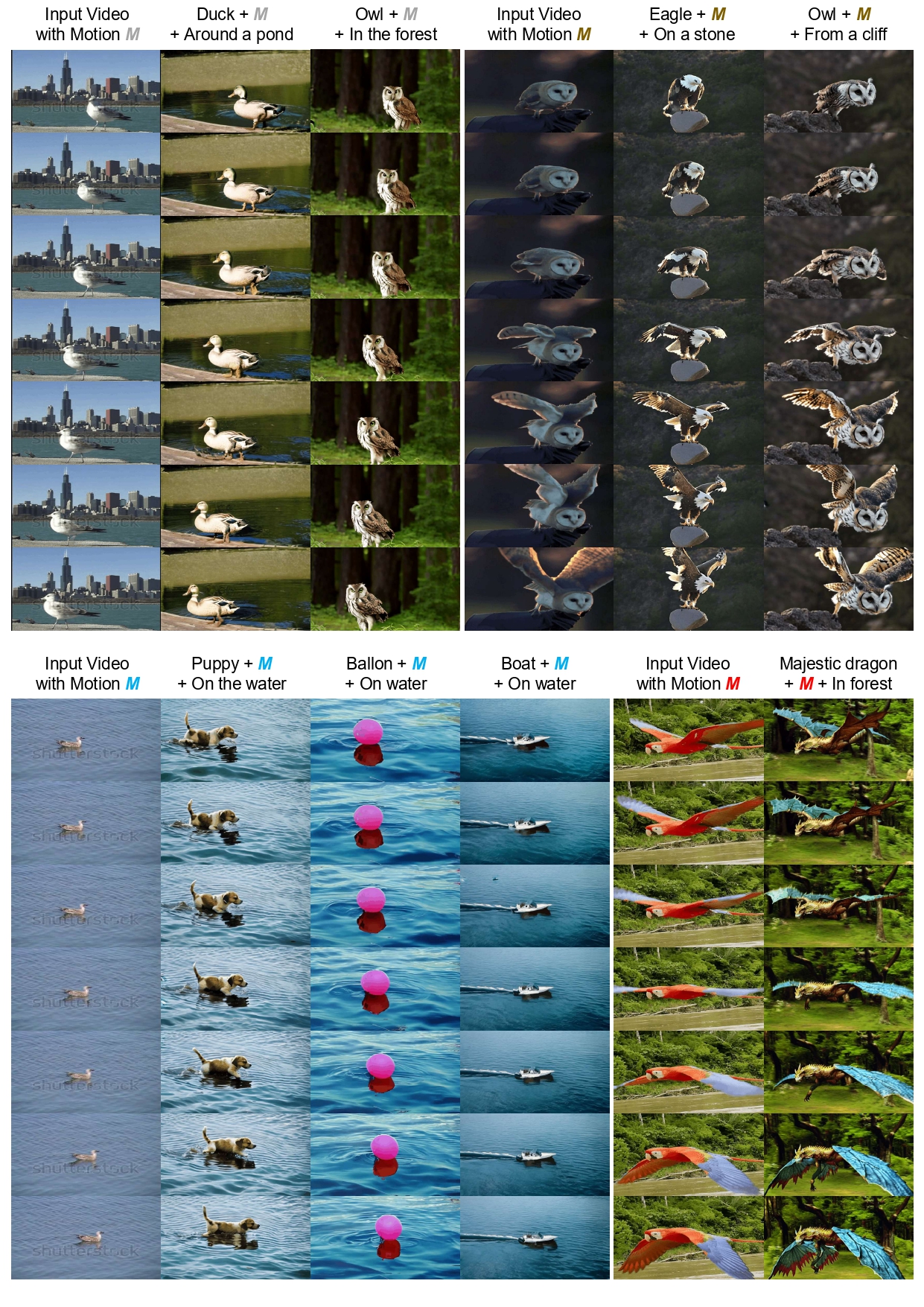}
\caption{Video Motion Customization results: Keyframes visualized.}
\label{fig:part3}
\end{figure*}

\begin{figure*}[htbp]
\centering
\includegraphics[width=0.9\textwidth]{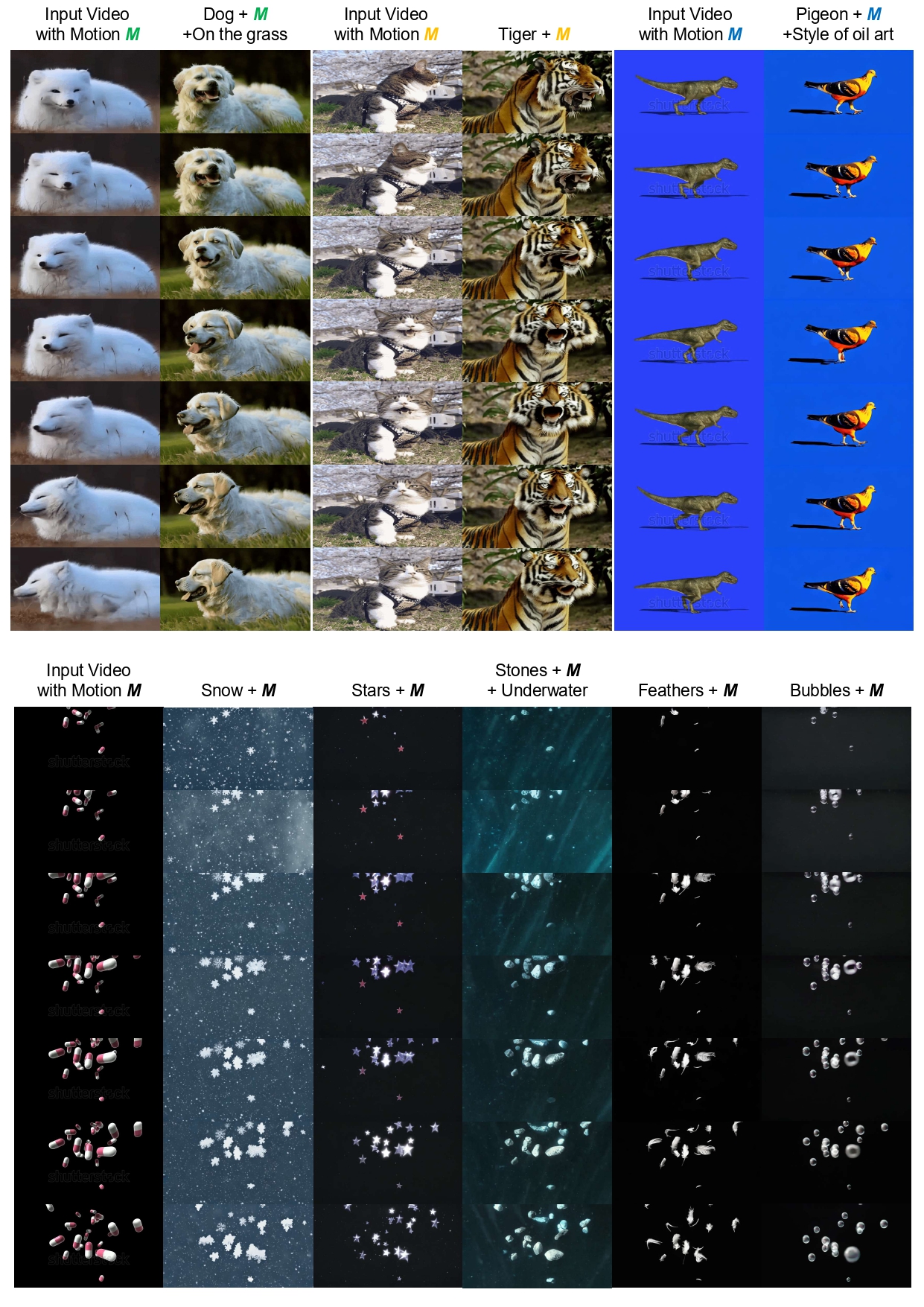}
\caption{Video Motion Customization results: Keyframes visualized.}
\label{fig:part4}
\end{figure*}

\begin{figure*}[htbp]
\centering
\includegraphics[width=0.9\textwidth]{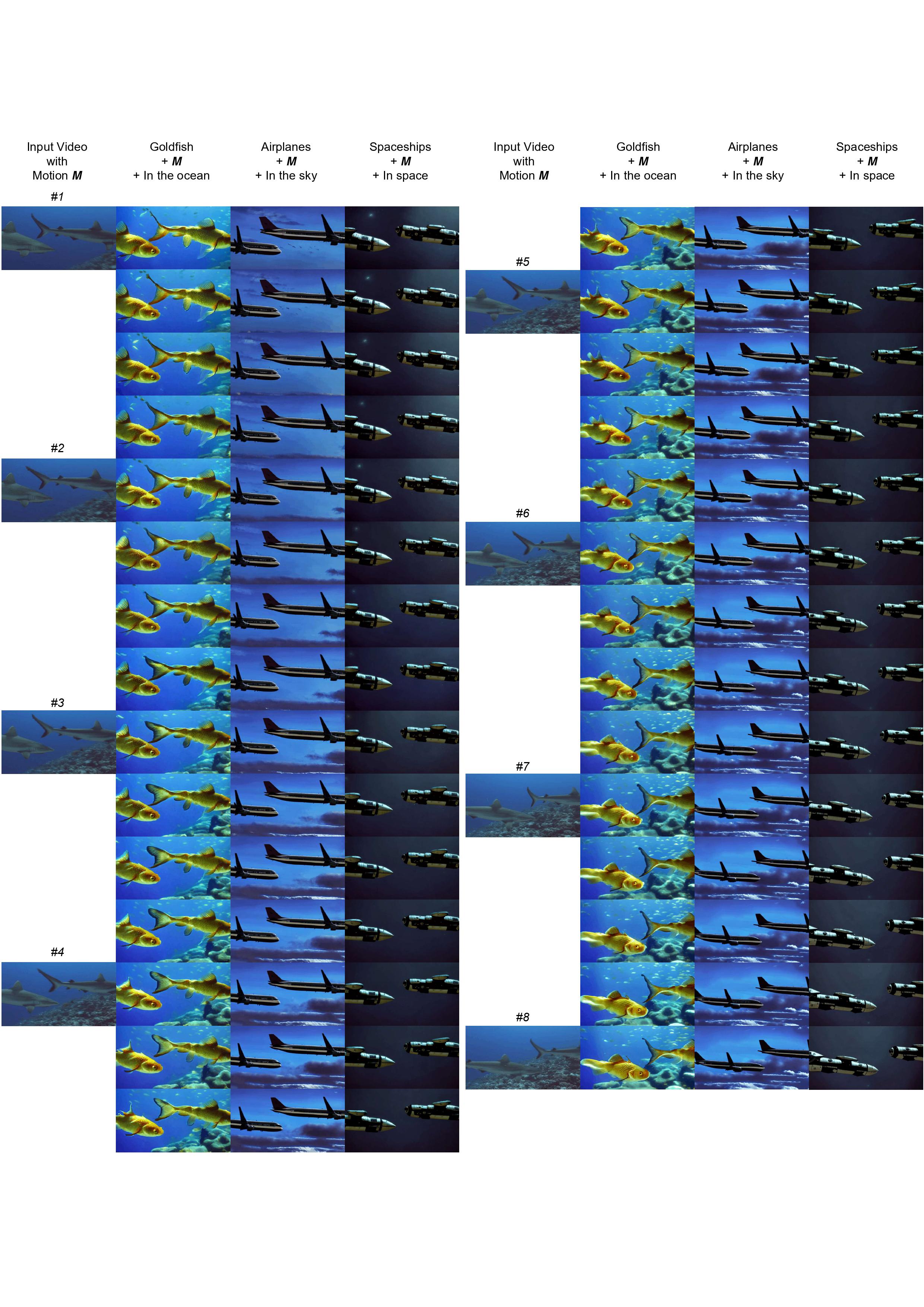}
\caption{
Full-frame results of Video Motion Customization: 
Text prompt ``Sharks are moving'' is used for training the keyframe generation UNet.
}
\label{fig:full1}
\end{figure*}

\begin{figure*}[htbp]
\centering
\includegraphics[width=0.9\textwidth]{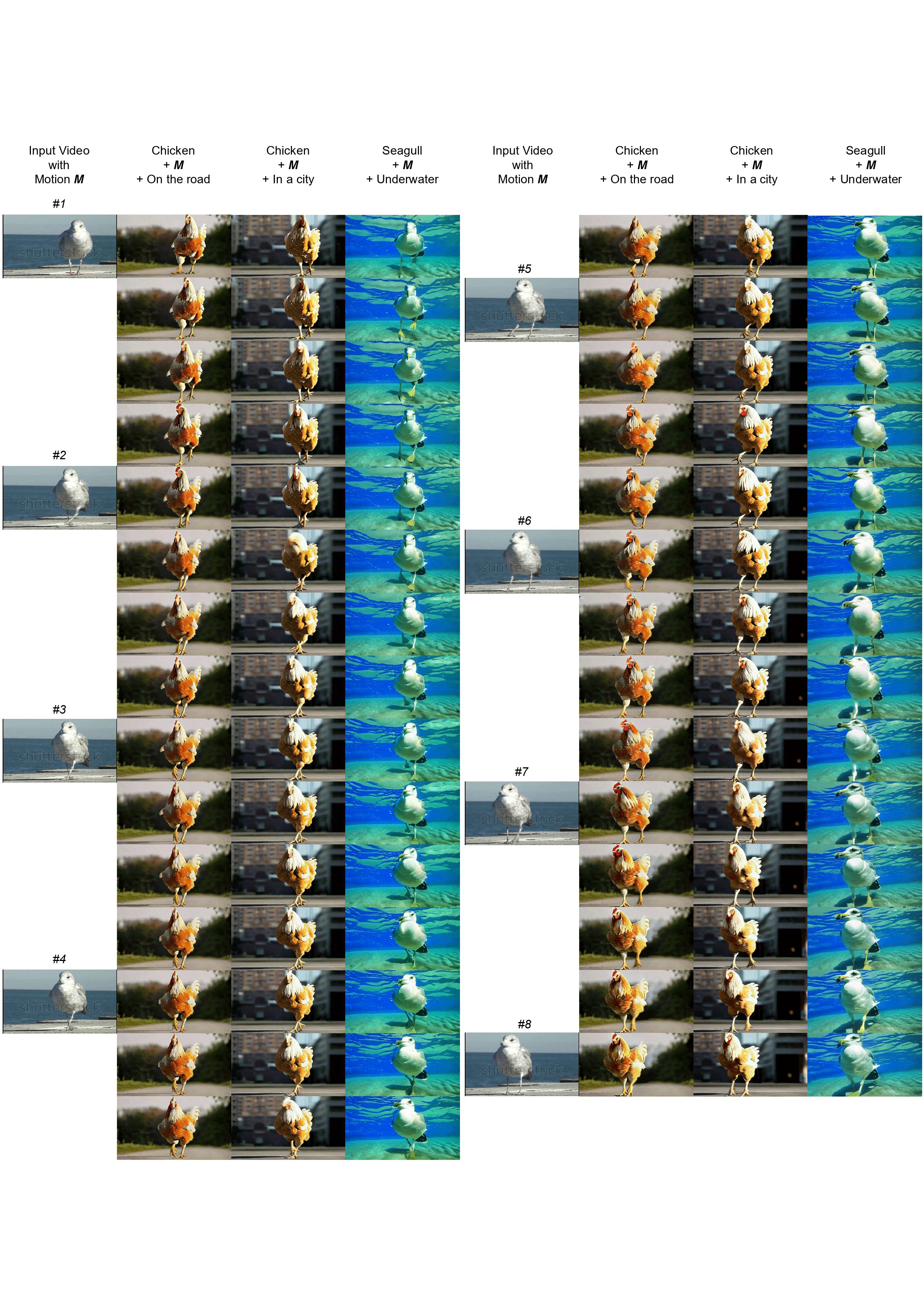}
\caption{
Full-frame results of Video Motion Customization: 
Text prompt ``A seagull is walking'' is used for training the keyframe generation UNet.
}
\label{fig:full2}
\end{figure*}

\begin{figure*}[htbp]
\centering
\includegraphics[width=0.9\textwidth]{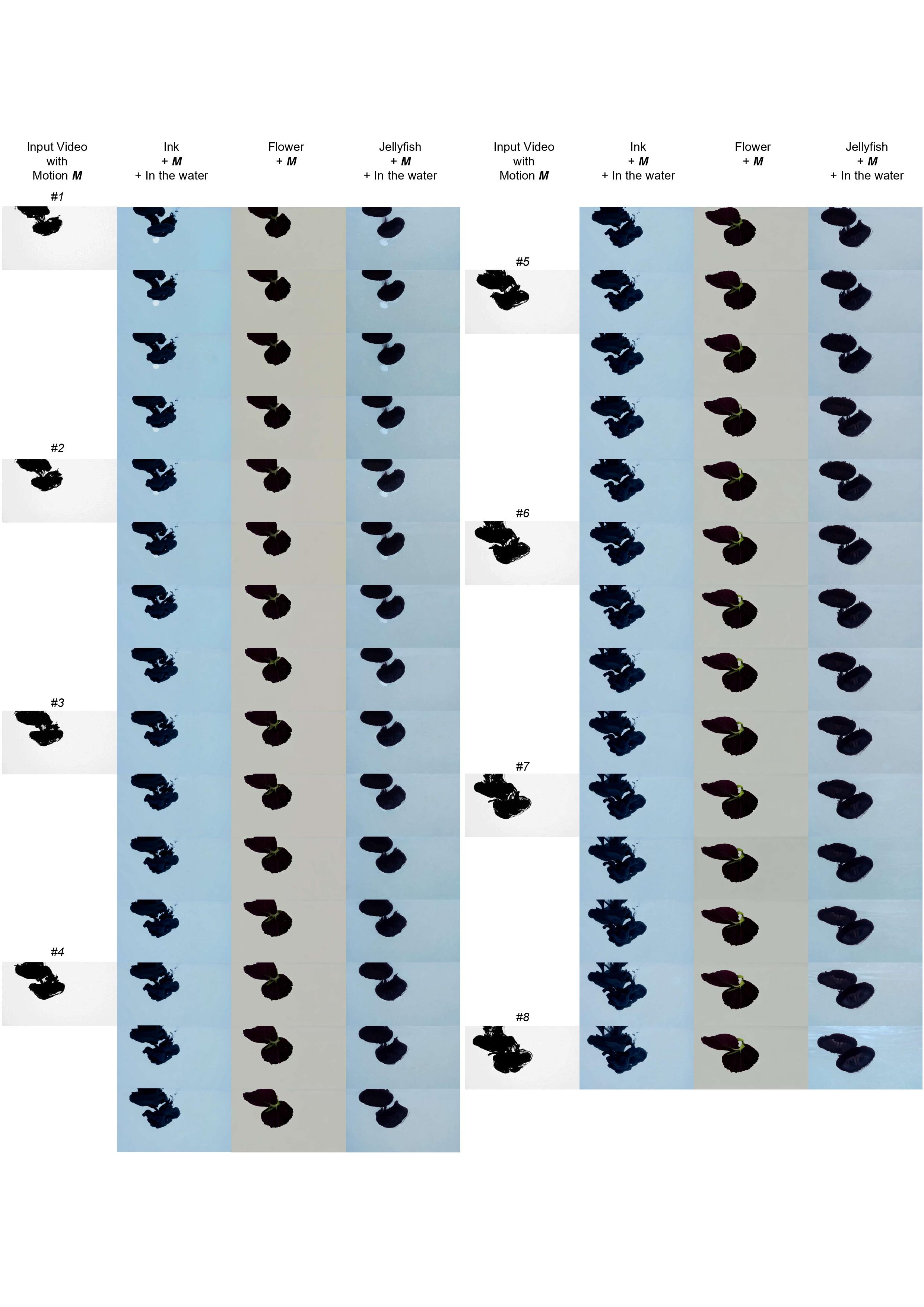}
\caption{
Full-frame results of Video Motion Customization: 
Text prompt ``Ink is spreading'' is used for training the keyframe generation UNet.
}
\label{fig:full3}
\end{figure*}

\begin{figure*}[htbp]
\centering
\includegraphics[width=0.9\textwidth]{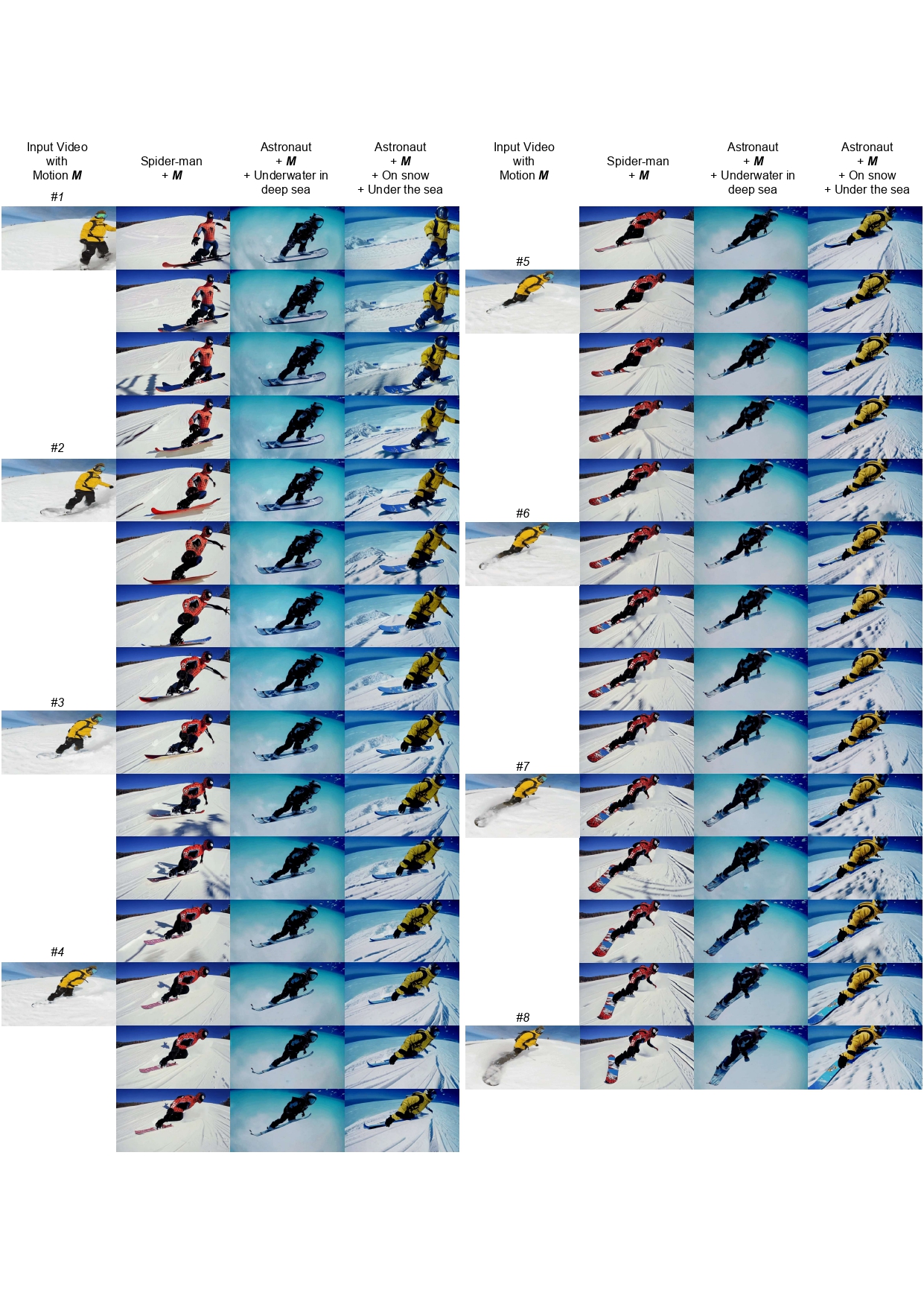}
\caption{
Full-frame results of Video Motion Customization: 
Text prompt ``A man is snowboarding'' is used for training the keyframe generation UNet.
}
\label{fig:full4}
\end{figure*}

\begin{figure*}[htbp]
\centering
\includegraphics[width=0.9\textwidth]{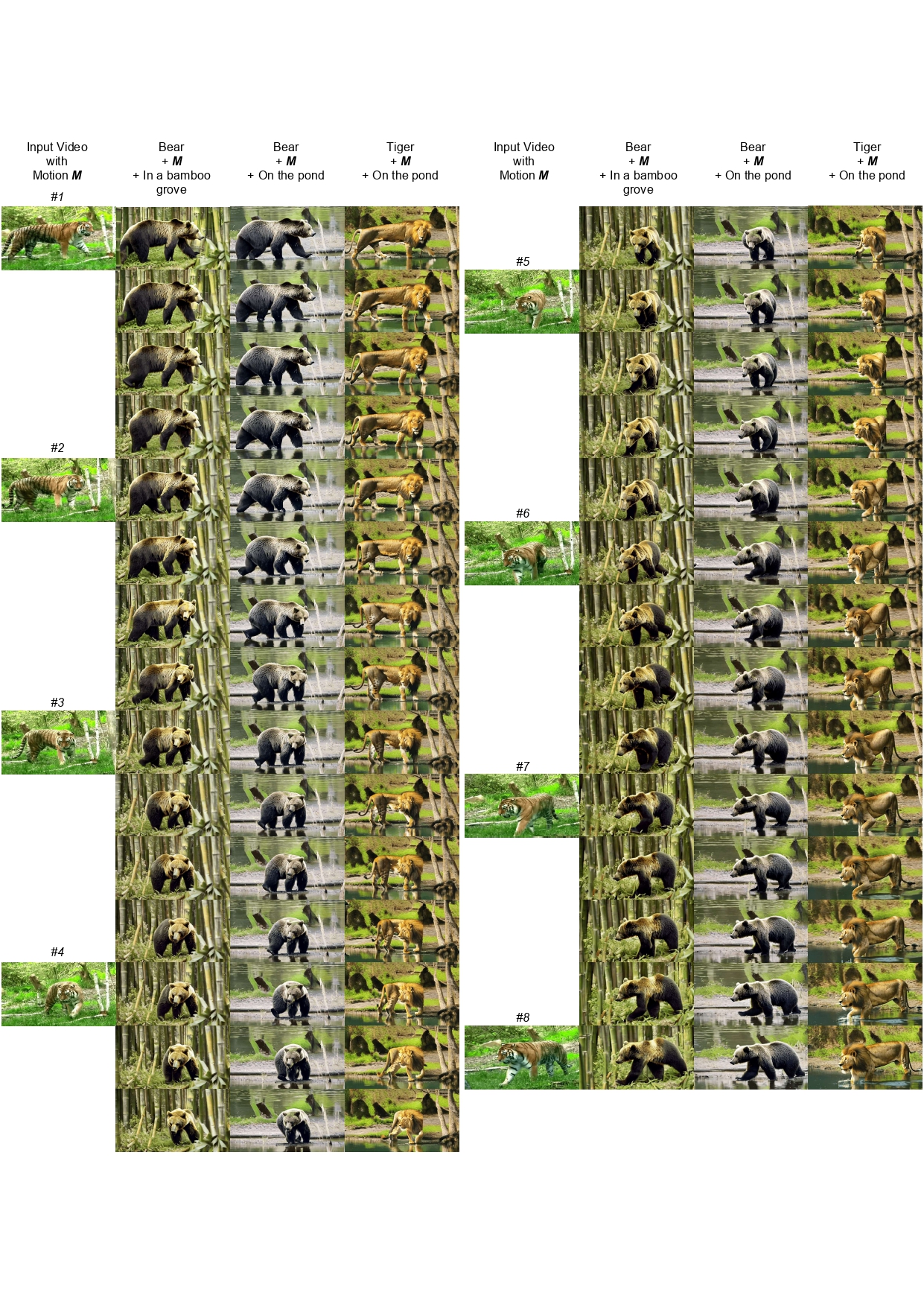}
\caption{
Full-frame results of Video Motion Customization: 
Text prompt ``A tiger is walking'' is used for training the keyframe generation UNet.
}
\label{fig:full5}
\end{figure*}

\quad This section is dedicated to presenting further results in motion customization. 
We display keyframes (7 out of the total 8 frames) from input videos in Figures \ref{fig:part1}, \ref{fig:part2}, \ref{fig:part3}, and \ref{fig:part4}, 
accompanied by various visual variations that maintain the essential motion patterns. 
Specifically, Figure \ref{fig:part1} showcases input videos featuring car movements. 
In Figure \ref{fig:part2}, we exhibit input videos capturing the dynamics of airplanes in flight and the blooming of a flower. 
Figure \ref{fig:part3} focuses on bird movements, including walking, taking off, floating, and flying. 
Lastly, Figure \ref{fig:part4}-top highlights input videos of mammals, 
while \ref{fig:part4}-bottom illustrates the motion of pills falling.
Moreover, for a comprehensive comparison between the motion in the input and generated videos, complete frames from these videos are presented in Figures \ref{fig:full1}, \ref{fig:full2}, \ref{fig:full3}, \ref{fig:full4}, and \ref{fig:full5}.
In each of these figures, the left columns show the 8-frame input video, while the adjacent three columns on the right exhibit 29 frames from the generated videos, replicating the same motion pattern.

\label{sec: additional results}
\clearpage

\end{document}